%% file: main.tex
\documentclass{article} 
\usepackage{iclr2027_conference,times}

\input{math_commands.tex}

\usepackage{hyperref}
\usepackage{url}

\usepackage{amssymb}
\usepackage{amsmath}
\usepackage{algorithm}
\usepackage{algpseudocode}
\usepackage{enumitem}
\usepackage{booktabs}
\usepackage[table]{xcolor}
\usepackage{cleveref}
\usepackage{float}
\usepackage{graphicx}

\newcommand{\methodname}{RESTA}
\newcommand{\bestresult}[1]{\cellcolor{blue!10}\textbf{#1}}

\usepackage[textsize=tiny]{todonotes}

\newtoggle{final}
\toggletrue{final}

\newcommand{\pw}[1]{\iftoggle{final}{#1}{{\color{blue} #1}}}

\title{\methodname: Time-Scaling Augmentation for Data-Efficient Robot Imitation Learning}

\author{
Yunpeng Jiang \\
Global College, Shanghai Jiao Tong University \\
Shanghai, China
\And
Jianshu Hu \\
Shanghai Institute of Physical Intelligence and Robotics (PAIR) \\
Shanghai, China
\And
Han Fang \\
Global College, Shanghai Jiao Tong University \\
Shanghai, China
\And
Zhiyuan Guan \\
Global College, Shanghai Jiao Tong University \\
Shanghai, China
\And
Paul Weng \\
Duke Kunshan University \\
Jiangsu, China
\And
Yutong Ban\thanks{Corresponding author: \texttt{yban@sjtu.edu.cn}} \\
Global College, Shanghai Jiao Tong University \\
Shanghai, China
}

\iclrfinalcopy 
\begin{document}

\maketitle
\fancyhead[]{}

\begin{abstract}
Robot manipulation policies trained through imitation learning commonly rely
on costly demonstrations recorded in one task direction and at one temporal
resolution. We propose \textbf{Re}versal and \textbf{S}tride-based
\textbf{T}rajectory \textbf{A}ugmentation (\methodname{}), which exploits
time-scaling data augmentation to improve data efficiency through
trajectory reversal and temporal resampling.
Trajectory-reversal augmentation reverses and relabels demonstrations between
paired tasks with opposite objectives, reusing recorded rollouts without
additional environment interaction. Because contact-rich manipulation is not
exactly reversible, we treat reversal as an approximate paired-task prior
rather than a physical symmetry. Temporal-stride augmentation varies
observation and action sampling strides and conditions the policy on the
resulting offsets, enabling fixed or randomized temporal schedules.
Experiments across 16 paired bimanual manipulation tasks show that trajectory
reversal is particularly effective in the low-data regime and that
temporal-stride augmentation improves success under randomized temporal
schedules. With 25 and 50 independently collected demonstrations per task,
\methodname{} achieves 78.4\% and 85.2\% average success,
respectively, matching or exceeding baselines trained with twice as many
demonstrations (77.3\% and 85.4\%). These results show that \methodname{}
can improve data efficiency in robot imitation learning.
\end{abstract}

\input{1_Introduction}

\input{2_Related_Works}

\input{3_Background}

\input{4_Methodology}

\input{5_Experimental_results}

\input{6_Conclusion}

\input{7_AI_statement}

\bibliography{main}
\bibliographystyle{iclr2027_conference}

\include{appendix}

\end{document}

%% file: math_commands.tex
\usepackage{amsmath,amsfonts,bm}

\def\eqref#1{equation~\ref{#1}}

\def\1{\bm{1}}

\DeclareMathAlphabet{\mathsfit}{\encodingdefault}{\sfdefault}{m}{sl}
\SetMathAlphabet{\mathsfit}{bold}{\encodingdefault}{\sfdefault}{bx}{n}



%% file: 1_Introduction.tex
\section{Introduction}
\label{sec:introduction}

Imitation learning offers a direct way to teach robots manipulation skills
\citep{zhao2023act,chi2023diffusionpolicy,ze2024dp3}, but \pw{requires} demonstrations that are costly to collect\pw{: each} trajectory \pw{needs} resetting the scene, executing the task, and
checking the result. 
\pw{Since a} demonstration also carries temporal structure \pw{(i.e., time direction and interval between observations and actions)}
\pw{, a} \pw{set of}
demonstration may \pw{thus} provide \pw{potential} supervision both across task
directions \pw{(e.g., open vs close door)} and across temporal resolutions \pw{(e.g., accelerating vs decelerating)}. 
This motivates our \pw{research}
question: \textit{Can temporal transformations of existing demonstrations
improve imitation learning across task directions and execution
schedules, without additional data collection?}

We use \emph{time scaling} to describe transformations of the temporal
coordinate used to construct policy-training windows. In physics, mathematics,
and computer simulation, time scaling refers to changing the ratio between
simulation time and real-world time: it can speed up or slow down a process,
or formally run its temporal coordinate backward. In our setting, the two
relevant properties are \emph{time reversal} and \emph{frequency change}.
Time reversal \citep{jiang2025timereversal,
qiao2026reversetoadvance,kim2026reverseplayback,barkley2023timereversal}
changes the direction of a sequence: for paired tasks such as opening and
closing an articulated object, a recorded opening
trajectory can provide potential supervision for closing when its observations
are reversed and actions are relabeled toward the next robot configuration.
This correspondence is approximate and depends on whether the reversed
transitions are realizable in the environment.
Frequency change, in contrast,
preserves temporal order while changing the \pw{time intervals} between observations or
action targets. 
It can be achieved in several ways, including changing action
persistence \citep{hu2024statenoveltyguidedactionpersistence}, using
multi-frequency action chunks \citep{zhang2026hipolicy}, or sparsely sampling
target sequences \citep{bai2026flash,park2025proleptic}. 
In this work, we use
positive temporal strides to change frequency: sampling a trajectory every
step defines the reference speed; increasing the stride samples more widely
separated states and corresponds intuitively to accelerating the robot's
motion, whereas decreasing it samples more densely and corresponds to
decelerating it.
Thus,
reversal changes the task direction \pw{of} a demonstration, while our
stride-based augmentation changes its temporal resolution.
\pw{During training, both can be naturally jointly applied.}

To exploit these two properties, we propose \textbf{Re}versal and \textbf{S}tride-based
\textbf{T}rajectory \textbf{A}ugmentation (\methodname), which combines paired-task trajectory
reversal with temporal-stride augmentation and temporal offset conditioning. 
\pw{In more details, r}eversal uses a trajectory from one task direction to provide
\pw{additional} supervision for its paired direction, reducing the need to collect
independent demonstrations for both. We form these examples by reversing the
observation sequence and relabeling each action toward the next robot
configuration. Temporal-stride augmentation samples histories and action
chunks at different positive strides, allowing the same demonstrations to
represent varied \pw{control frequencies}. 
Because the policy must distinguish these \pw{frequencies}, we provide the
temporal offset, which contains both observation and action offsets, through a lightweight embedding.
This conditioning can be adapted to different policy architectures;
at inference, different offset schedules set the \pw{time intervals} between predicted
targets, allowing the policy to operate at higher or lower effective
frequencies.

\begin{figure*}[t]
    \centering
    \includegraphics[width=\textwidth]{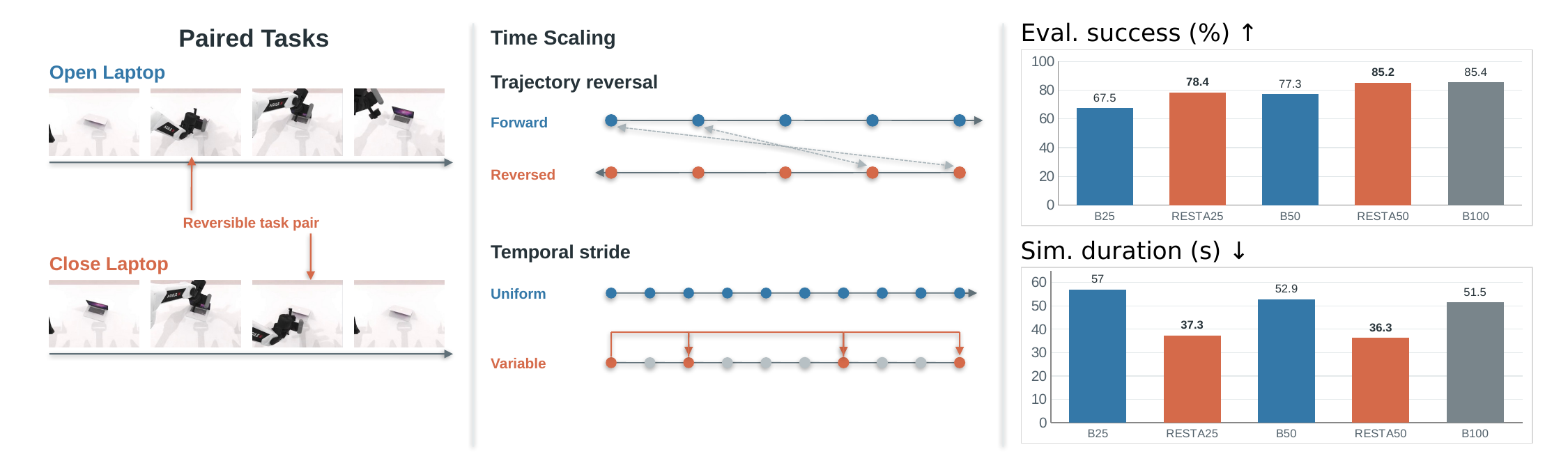}
    \caption{Overview and aggregate results for \methodname{}. Left:
    Open Laptop and Close Laptop trajectories illustrate a reversible task
    pair. Middle: time scaling is illustrated through trajectory reversal,
    which connects representative states across forward and reversed time,
    and temporal-stride augmentation, which samples a shared timeline
    uniformly or variably. Right: bars report mean evaluation success and
    approximate simulated duration over the 16-task suite; higher success and
    lower duration are preferred. $B_n$ denotes DP3 trained with $n$ native
    demonstrations per task, and $\mathrm{\methodname}_n$ adds trajectory-reversal
    and temporal-stride augmentation.}
    \label{fig:teaser}
\end{figure*}

\paragraph{Contributions}
Our contributions can be summarized as follows (\Cref{fig:teaser}):
\begin{enumerate}[label=(\roman*),wide]
    \item We formulate the imitation-learning training-window problem and
    its time-scaling view, with each window pairing an observation history
    with future action targets. This gives a common view of trajectory
    reversal and positive temporal resampling, exposing two ways to reuse
    temporal structure in demonstrations
    (\Cref{sec:method-problem-formulation,sec:method-time-scaling}).
    \item We introduce paired-task trajectory reversal as a data augmentation
    (\Cref{sec:method-reversal}) that creates candidate behavior-cloning
    examples by reversing observations and relabeling absolute
    next-configuration targets, without additional environment interaction.
    \item We introduce temporal-stride augmentation with observation and action
    offset conditioning (\Cref{sec:method-stride}) to vary the \pw{frequency in}
    demonstration windows and tell the policy \pw{at} which \pw{time intervals} it should
    represent and predict.
    \item We evaluate \methodname{} on sixteen paired bimanual tasks
    (\Cref{sec:exp-main}). At matched data
    budgets, it achieves higher average success and requires fewer simulator
    physics steps per successful episode than the baseline, indicating 
    more reliable task completion and shorter simulated execution time. We
    also compare against larger-data baselines and analyze the contribution
    of each augmentation.
\end{enumerate}

%% file: 2_Related_Works.tex
\section{Related Work}
\label{sec:related-works}

We organize \pw{the related work} around two directions: action-chunk imitation
learning and temporal transformations. Within temporal transformations, we
distinguish time-reversal methods from approaches that change action
frequency.
Due to page limit, discussions of action-chunk imitation policies and temporal-frequency
methods appear in \Cref{sec:app-related-work}.


\paragraph{Temporal Symmetry for Robotics Control}

Time Reversal Deep Reinforcement Learning (TR-DRL)
\pw{\citep{jiang2025timereversal}} filters dynamics-consistent reversible
transitions,
augments replay data, and shapes rewards using reversal;
Time-Symmetric Data Augmentation \pw{\citep{barkley2023timereversal}} also
constructs reverse-time transitions for off-policy RL and studies failures
from irreversible dynamics. Other uses include symmetry-guided quadruped gait
transitions \citep{ding2024quadrupedalsymmetry}, self-supervised manipulation
planning \citep{nair2018trass}, and reverse curricula that alter the state
distribution \pw{\citep{florensa2017reversecurriculum,tao2024rfcl}}. We instead
construct behavior-cloning examples from paired manipulation demonstrations.
\pw{Recent work also studies demonstration reversal.} Reverse Playback Motion
\pw{\citep{kim2026reverseplayback}} has an operator execute a motion from its
goal state, then reconstructs a forward trajectory by reversing recorded
end-effector poses. We reverse observations and relabel targets in an existing
expert dataset, without additional execution or collection. Reverse to Advance
\pw{\citep{qiao2026reversetoadvance}} iteratively uses environment-collected
easy- and hard-task rollouts, retaining segments with kinematic and
critic-guided advantage filters; a learned dynamics filter appears only in a
comparison baseline. Our method uses fixed expert trajectories without policy
rollouts, resets, or new data, and also augments temporal strides and
observation offsets.

%% file: 3_Background.tex
\section{Background}
\label{sec:background}

The background covers the foundations of action-chunk imitation learning:
the Markov decision-process formulation and offline imitation learning
(\Cref{sec:bg-mdp,sec:bg-imitation}), followed by 3D Diffusion Policy \pw{(DP3)} as the
action-chunk policy backbone (\Cref{sec:bg-dp3}).

\subsection{Markov Decision Processes}
\label{sec:bg-mdp}

We use a finite-horizon Markov decision process (MDP) to model robot
manipulation. It is specified by
$\mathcal{M}=(\mathcal{S},\mathcal{A},P,\rho,r,T)$, with full
robot--environment state space $\mathcal{S}$, action space
$\mathcal{A}$, transition kernel $P$, initial-state distribution $\rho$,
reward $r$, and horizon $T$. At the start of each episode, the initial
state is sampled as $s_0\sim\rho$, where $\rho$ is the distribution over
robot--environment states induced by the reset procedure. At step $t$, state
$s_t\in\mathcal{S}$ evolves
as $s_{t+1}\sim P(\cdot\mid s_t,a_t)$ under action $a_t\in\mathcal{A}$.
Actions specify absolute joint-position and gripper targets. A supplied
low-level controller converts each target into the robot commands executed
over the next transition, so its tracking behavior is included in the
transition kernel $P$.
The objective is to reach the success set
$\mathcal{G}\subseteq\mathcal{S}$ within $T$ steps. Episodes terminate at
first success or the horizon, with reward one for success and zero otherwise \pw{(provided by reward function $r$)}.
The policy uses sensory observations $o_t$ in a length-$L$ history
$h_t=(o_{t-L+1},\ldots,o_t)$ instead of full states. 

\subsection{Imitation Learning}
\label{sec:bg-imitation}

Offline imitation learning fits a policy to $N$ expert trajectories without
rewards or a transition model. Let
$\mathcal{D}=\{\tau^{(n)}\}_{n=1}^{N}$, where
$\tau^{(n)}=(o_0,a_0,\ldots,a_{T_n-1},o_{T_n})$ has $T_n$ transitions.
At timestep $t$, the policy receives the history
$h_t=(o_{t-L+1},\ldots,o_t)$ and predicts an $H$-action chunk. A training
window is the pair $(h_t,A_t^*)$, where
$A_t^*=(a_t^*,\ldots,a_{t+H-1}^*)$; valid windows satisfy
$t\geq L-1$ and $t+H-1<T_n$.

At deployment, the history has length $L$; before $L$ observations are
available, the earliest observation is repeated to left-pad it, and then the
most recent $L$ observations are retained. The policy executes part of each
predicted chunk, collects new observations, and replans. Behavior cloning (BC)
fits the policy to these expert windows with
the objective
\begin{equation}
    \mathcal{L}_{\mathrm{IL}}(\theta)
    =\mathbb{E}_{(h_t,A_t^*)\sim\mathcal{D}}
    \left[\ell\!\left(\pi_\theta(\cdot\mid h_t),A_t^*\right)\right],
    \label{eq:bg-imitation-objective}
\end{equation}
where $\pi_\theta(\cdot\mid h_t)$ is the predictive distribution and $\ell$
measures its disagreement with the expert chunk and the expectation ranges over
valid windows. Here, windows use consecutive trajectory samples; our method later
varies their temporal offsets in \Cref{sec:method-time-scaling}.

\subsection{3D Diffusion Policy \pw{(DP3)}}
\label{sec:bg-dp3}

\pw{DP3 \citep{ze2024dp3}} represents $\pi_\theta(A_t\mid h_t)$ through
conditional denoising. An encoder $e_\phi$ with parameters $\phi$ combines
point-cloud features and proprioception across $h_t$ into $c_t=e_\phi(h_t)$.
In our action-chunk setting, the policy predicts $H$ absolute joint-position
and gripper targets from this observation history.
Let
$\psi(o_t)$ extract the recorded joint positions and gripper configuration in
the controller's target representation. For consecutive samples, the
supervised \pw{action} chunk is $(\psi(o_{t+1}),\ldots,\psi(o_{t+H}))$, with
$t+H\leq T$. A supplied low-level controller drives the robot toward each
predicted target; it is not learned, and its tracking behavior is included in
the transition model. The objective is to learn action-chunk predictions
from demonstrations that enable successful closed-loop task execution.
For a normalized expert chunk $x_0$, diffusion step $n$ produces
\begin{equation}
    x_n=\sqrt{\bar{\alpha}_n}\,x_0
        +\sqrt{1-\bar{\alpha}_n}\,\epsilon,
    \qquad \epsilon\sim\mathcal{N}(0,I),
    \label{eq:bg-forward-diffusion}
\end{equation}
where $x_n$ is the noisy chunk, $\bar{\alpha}_n$ is cumulative signal
retention, $\epsilon$ is Gaussian noise, and $I$ is the identity covariance.
Training learns denoising conditioned on $c_t$ and $n$ using the objective
\begin{equation}
    \mathcal{L}_{\mathrm{DP3}}(\theta;x_0,c_t)
    =\mathbb{E}_{n,\epsilon}
      \left[
        \left\|\epsilon-\epsilon_\theta(x_n,c_t,n)\right\|_2^2
      \right],
    \label{eq:bg-dp3-objective}
\end{equation}
where $\epsilon_\theta$ predicts the noise added at diffusion step $n$.
Inference iteratively converts noise into an action chunk.

%% file: 4_Methodology.tex
\section{\methodname{}}
\label{sec:methodology}

We learn action-chunk policies from expert demonstrations using two
augmentations based on the time-scaling framework introduced below.
After defining the control problem and paired-task setting in
\Cref{sec:method-problem-formulation}, we introduce the time-scaling
framework in
\Cref{sec:method-time-scaling}, then construct reversed demonstrations
to share supervision between task directions in \Cref{sec:method-reversal}.
We then vary the temporal offset of training windows and condition the
policy on that offset in \Cref{sec:method-stride}.
\Cref{sec:method-combined} brings these steps together for training and
inference.

\subsection{Problem Formulation}
\label{sec:method-problem-formulation}

We consider learning robot control from a fixed dataset $\mathcal{D}$ of
expert demonstrations. The action-chunk policy and its target
representation are defined in the DP3 background; here we specify the
paired-task setting used for trajectory reversal.
The tasks are organized into pairs with expert datasets $\mathcal{D}^{+}$
and $\mathcal{D}^{-}$, such as opening and closing an articulated object.
We construct reverse-task supervision by reversing recorded trajectories and
relabeling each action toward the next configuration in the reversed sequence.
This is an offline augmentation procedure, not an assertion that every
reversed transition is physically realizable.

\subsection{Time Scaling}
\label{sec:method-time-scaling}

Time scaling changes how a demonstrated trajectory is indexed to construct
a training window $(h_t,A_t)$, a finite portion of the trajectory used to
form one training example. These windows appear in
\Cref{eq:bg-imitation-objective}. Let $\xi\in[0,T]$ denote the source
coordinate along the original trajectory, with $T$ its final
source-coordinate value.
For a window with transformed sampling instants
$0=v_0<v_1<\cdots<v_N=\mathcal{V}$, where $\mathcal{V}$ is the transformed
window endpoint, define the mapping $\xi(v)$ from transformed position to
source-coordinate location by
\begin{equation}
    \xi(v_0)=\xi_0, \qquad
    \xi(v_k)=\xi(v_{k-1})+\lambda_k(v_k-v_{k-1}),
    \qquad \lambda_k\ne 0,
    \label{eq:method-time-scaling}
\end{equation}
where $\xi_0$ is the source-coordinate location at the beginning of the
window. The local scaling factor $\lambda_k$ may vary across steps. The
window is valid when $0\leq\xi(v_k)\leq T$ for every $k$. The sign of
$\lambda_k$ determines the local traversal direction, while $|\lambda_k|$
determines the source-coordinate spacing for that step. Constant scaling is
the special case $\lambda_k=\lambda$ for all $k$.
For positive scaling, a larger value of $\lambda_k$ advances farther along
the demonstrated trajectory per unit transformed time and therefore
corresponds to accelerating along the trajectory; a smaller positive value
corresponds to decelerating. When the scale varies across a window, the
low-level controller must track these changing effective rates.

Reversal uses the constant schedule $\lambda_k=-1$ for all $k$, with
$\xi_0=T$ and
$\mathcal{V}=T$, so that $\xi(v)=T-v$.
Let $\mathbf{s}(\xi)$ denote the full state trajectory. Physical reversal then
gives $\mathbf{s}^{\mathrm{rev}}(v)=\mathcal{R}\bigl(\mathbf{s}(\xi(v))\bigr)
=\mathcal{R}\bigl(\mathbf{s}(T-v)\bigr)$, where $\mathcal{R}$ preserves
positions and negates velocities.
For the absolute
targets in \Cref{sec:bg-mdp}, reversed demonstrations require actions
relabeled toward preceding source configurations, rather than simply
reversed action arrays. This provides reverse-task supervision for paired
tasks such as opening and closing, but contact, friction, delays, and grasp or
release events can make some reversed transitions infeasible.

Positive scaling preserves order. For uniform recording frequency $f_0$
and period $\Delta t_0=1/f_0$, sampling $v$ every $\Delta t_0$ with
$\lambda=g+1$ skips $g\in\{0,1,2,\ldots\}$ source samples. The resulting
source-time interval and sampling frequency are
\begin{equation}
    \Delta t_g=(g+1)\Delta t_0,
    \qquad
    f_g=\frac{f_0}{g+1}.
    \label{eq:method-stride-frequency}
\end{equation}
Variable strides give a piecewise time transformation.
Thus, reversal changes the task direction represented by supervision, while
resampling changes its temporal resolution.

\paragraph{Idealized transition reversibility.}
Using the transition model in \Cref{sec:bg-mdp}, write $P^{+}$ and
$P^{-}$ for the paired task directions. Following the definition of full
time-reversal symmetry in \citet{jiang2025timereversal}, an idealized
reversibility condition would require that for each transition
$(s_t,a_t,s_{t+1})$, there exists
a reverse action $\hat{a}_t\in\mathcal{A}$ satisfying
\begin{equation}
    P^{+}(s_{t+1}\mid s_t,a_t)
    =P^{-}(\mathcal{R}s_t\mid
           \mathcal{R}s_{t+1},\hat{a}_t).
    \label{eq:transition-reversibility}
\end{equation}
This equality describes an idealized condition for exact transition
reversibility; our method does not assume that it holds for every transition
in the demonstrations. We nevertheless use reversed trajectories and their relabeled
targets directly for training, without segment-level feasibility filtering.
This simple augmentation is effective in our experiments
(\Cref{sec:exp-main,sec:app-data-efficiency}), while filtering infeasible
segments is a promising direction for future work.

\paragraph{Strided target prediction.}
Building on the DP3 action-chunk formulation in \Cref{sec:bg-dp3}, we now allow
the observation history and target samples within a chunk to use positive
adjacent offsets rather than the default consecutive spacing. 
Consider a discrete demonstration indexed by $t=0,\ldots,T$, where $T$ is
the final trajectory index.
For an anchor $t$,
we randomly sample observation indices $0\leq i_1<\cdots<i_L=t$ and target
indices $t<j_1<\cdots<j_H\leq T-1$ according to the selected observation and
action strides. Consecutive sampling is the special case
$i_\ell=t-L+\ell$ and $j_k=t+k$. The resulting observation history and action
targets are $(o_{i_1},\ldots,o_{i_L})$ and
$(a_{j_1},\ldots,a_{j_H})=(\psi(o_{j_1+1}),\ldots,\psi(o_{j_H+1}))$,
respectively, where each action target is the configuration of the next
observation.


In the next two subsections, we explain how to exploit time scaling in imitation learning.

\subsection{Trajectory Reversal Augmentation}
\label{sec:method-reversal}

Trajectory reversal augments each task's expert dataset with demonstrations
from its paired task using the offline construction in
\Cref{sec:method-problem-formulation}.
The construction reverses the observation order and relabels each action
to target the next configuration in the reversed sequence.

\paragraph{Reversed demonstrations.}
Given a source trajectory
$\tau=(o_0,a_0,\ldots,a_{T-1},o_T)$ with $T$ transitions, we reverse
the observation order:
\begin{equation}
    \tilde{o}_t=o_{T-t}, \qquad t=0,\ldots,T.
    \label{eq:reverse-state-sequence}
\end{equation}
This applies the reversal case $\lambda=-1$ of \Cref{eq:method-time-scaling}
to the recorded point clouds and robot configurations.

Using the configuration extractor $\psi$ defined in
\Cref{sec:bg-dp3}, the action at each reversed
observation targets the next robot configuration in the reversed sequence:
\begin{equation}
    \tilde{a}_t=\psi(\tilde{o}_{t+1})
               =\psi(o_{T-t-1}), \qquad t=0,\ldots,T-1.
    \label{eq:reverse-action-label}
\end{equation}
The targets are thus obtained directly from recorded proprioception.

Let $\operatorname{Rev}$ denote observation reversal together with this
action relabeling. Applying it in both directions \pw{yields two augmented datasets, which can be used to train for their respective tasks:}
\begin{equation}
    \widetilde{\mathcal{D}}^{+}
    =\mathcal{D}^{+}\cup\operatorname{Rev}(\mathcal{D}^{-}),
    \qquad
    \widetilde{\mathcal{D}}^{-}
    =\mathcal{D}^{-}\cup\operatorname{Rev}(\mathcal{D}^{+}).
    \label{eq:augmented-paired-datasets}
\end{equation}
\pw{The two} datasets reuse the recorded demonstrations without \pw{extra}
environment interaction. We next describe how to sample training windows
from these trajectories at different temporal spacings.

\subsection{Temporal-Stride Augmentation}
\label{sec:method-stride}

For either an original or a reversed trajectory, consecutive training
samples correspond to $g=0$ in \Cref{eq:method-stride-frequency}.
Temporal-stride augmentation also samples windows with $g\geq 1$, so that
the policy learns to predict chunks of more widely spaced joint targets.
We apply the positive time scaling in \Cref{sec:method-time-scaling} to vary
both observation and action spacing while keeping the history length $L$
and action horizon $H$ fixed.

\paragraph{Training windows.}
Let $\mathcal{B}$ be the set of \pw{allowed} positive integer strides. For each
training window anchored at index $t$, we independently and randomly sample
the observation and action offset vectors. Let
$\boldsymbol{\delta}^{o}=(\delta^{o}_1,\ldots,\delta^{o}_{L-1})\in
\mathcal{B}^{L-1}$ and
$\boldsymbol{\delta}^{a}=(\delta^{a}_1,\ldots,\delta^{a}_H)\in
\mathcal{B}^{H}$. These are adjacent offsets: each entry is the gap
between neighboring samples, with the first action gap measured from the
anchor. They determine the time indices through
\begin{equation}
    i_L=t,\enspace i_\ell=i_{\ell+1}-\delta^{o}_\ell
    \enspace(\ell=1,\ldots,L-1),
    \quad j_0=t,\enspace j_k=j_{k-1}+\delta^{a}_k
    \enspace(k=1,\ldots,H),
    \label{eq:stride-index-relations}
\end{equation}
Since all adjacent offsets are positive, these indices satisfy
\begin{equation}
    0\leq i_1<\cdots<i_L=t=j_0<j_1<\cdots<j_H\leq T-1.
    \label{eq:stride-index-order}
\end{equation}
Thus,
$\boldsymbol{\delta}^{o}$ and $\boldsymbol{\delta}^{a}$ directly specify
the adjacent temporal spacing of the observation history and action chunk.
Writing $\eta=[\boldsymbol{\delta}^{o},\boldsymbol{\delta}^{a}]$ for
this adjacent-offset schedule, we construct
\begin{equation}
    h_t^{\eta}=(o_{i_1},\ldots,o_{i_L}),
    \qquad
    A_t^{\eta}=(a_{j_1},\ldots,a_{j_H}).
    \label{eq:temporal-window}
\end{equation}
These offsets describe spacing in the recorded sequence. For uniformly
timed recordings, their relation to sampling frequency follows
\Cref{eq:method-stride-frequency}.

\paragraph{Temporal-offset conditioning.}
Temporal-offset conditioning consists of two offset vectors:
\pw{the adjacent spacing $\boldsymbol{\delta}^{o}$ in the observation
history and the adjacent spacing $\boldsymbol{\delta}^{a}$ in the
future action chunk.}
We collect them in the schedule
$\eta=[\boldsymbol{\delta}^{o},\boldsymbol{\delta}^{a}]$. Since these two
offset vectors are sampled independently, the
observation history alone does not determine the temporal spacing of the
action targets. We therefore provide the full schedule $\eta$ to the policy
by augmenting the DP3 condition $e_\phi(h_t^{\eta})$ with a learned embedding
of $\eta$.
Let $g_\omega$ denote a learnable function with parameters
$\omega$ that encodes the normalized adjacent observation and action offsets.
The resulting condition and policy \pw{become:}
\begin{equation}
    c_t^{\eta}
    =\left[e_\phi(h_t^{\eta}),g_\omega(\eta)\right],
    \qquad
    \pi_\theta(A_t^{\eta}\mid h_t^{\eta},\eta),
    \label{eq:temporal-condition}
\end{equation}
where brackets denote concatenation and $\theta$ collects the trainable
policy parameters.

\paragraph{Inference schedules.}
Because the policy is conditioned on temporal offsets, inference must specify
the spacing of the future targets before each prediction. At replanning cycle
$q$, the scheduling rule $\sigma$ selects
$\boldsymbol{\delta}^{a}_q\in\mathcal{B}^H$ according to a fixed,
randomized, or context-dependent strategy. If $t_q$
is the current anchor, the target indices satisfy
\begin{equation}
    j_{q,0}=t_q,
    \qquad
    j_{q,k}=j_{q,k-1}+\delta^a_{q,k},
    \qquad k=1,\ldots,H.
    \label{eq:inference-action-schedule}
\end{equation}
For $q=0$, there is no previous action chunk, so the history is initialized
with configured observation offsets $\boldsymbol{\delta}^{o}_0
=\boldsymbol{\delta}^{o,\mathrm{init}}$ (consecutive observations correspond to
$\boldsymbol{\delta}^{o,\mathrm{init}}=(1,\ldots,1)$), and the initial
schedule is $\eta_0=[\boldsymbol{\delta}^{o,\mathrm{init}},
\boldsymbol{\delta}^{a}_0]$. For $q\geq 1$, the
current observation offsets are determined by the spacing of the previous
action chunk. Although observations arrive one at a time, the retained
history represents the gaps induced by the last chunk, so
\begin{equation}
    \boldsymbol{\delta}^{o}_q
    =\left(\delta^{a}_{q-1,H-L+2},\ldots,
      \delta^{a}_{q-1,H}\right).
    \label{eq:inference-observation-offsets}
\end{equation}
For example, with $L=3$, $H=6$, and
$\boldsymbol{\delta}^{a}_{q-1}=(2,1,3,2,1,3)$, the transition from cycle
$q-1$ to cycle $q$ gives $\boldsymbol{\delta}^{o}_q=(1,3)$. The schedule
$\eta_q=[\boldsymbol{\delta}^{o}_q,\boldsymbol{\delta}^{a}_q]$ then conditions
the prediction via \Cref{eq:temporal-condition}. We study the choice of
$\sigma$ in the action-offset ablation in \Cref{sec:exp-ablation}.

\subsection{\methodname{} Algorithm}
\label{sec:method-combined}

The two augmentations enter training in sequence. We first form
$\widetilde{\mathcal{D}}^{+}$ and $\widetilde{\mathcal{D}}^{-}$ using
\Cref{eq:augmented-paired-datasets}. We then sample temporal schedules
when constructing windows from these datasets, so both original and
reversed demonstrations provide supervision at different spacings.

\begin{figure*}[t]
    \centering
    \includegraphics[width=\textwidth]{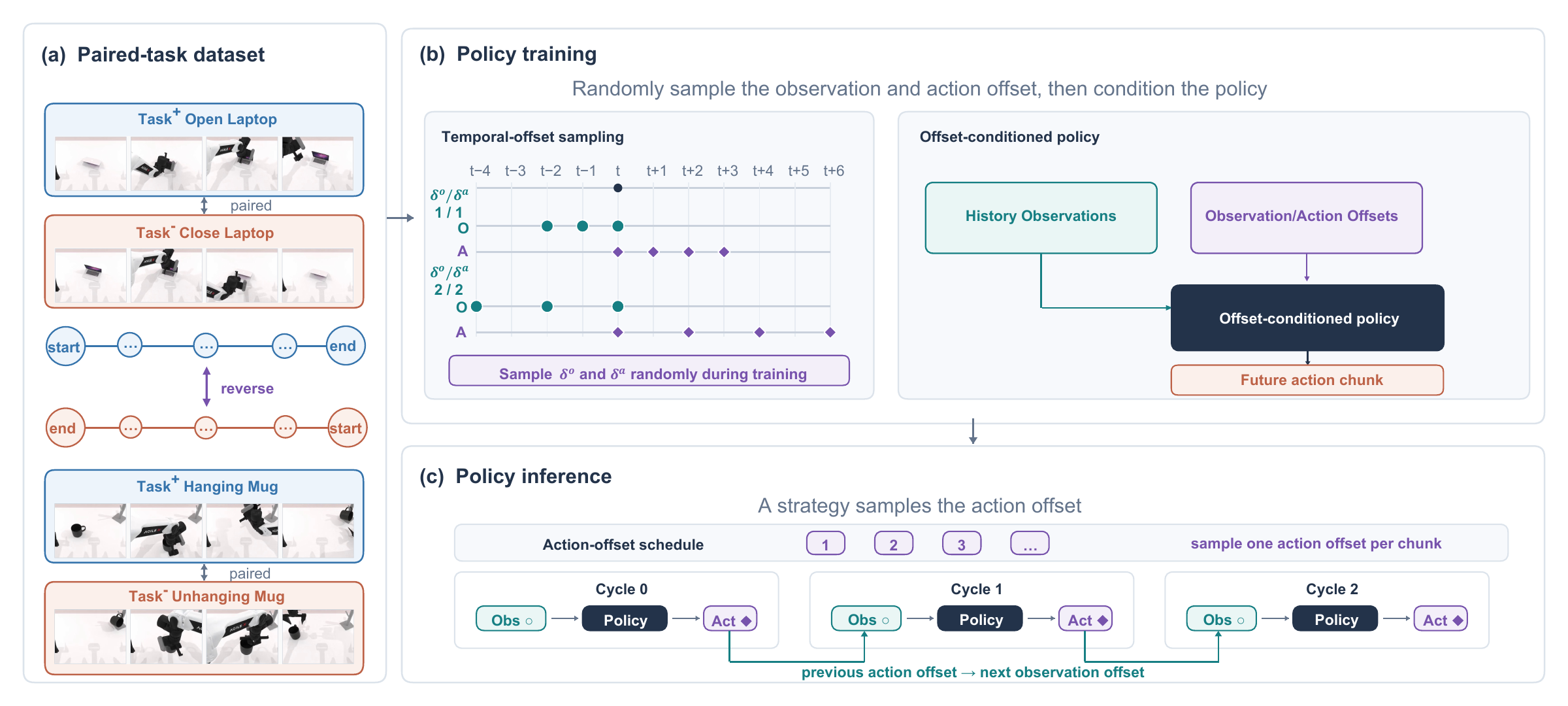}
    \caption{Overview of \methodname{}. (a): paired task directions provide
    demonstrations that are reversed and relabeled to construct augmented
    datasets. (b): during training, sampled temporal offsets determine the
    observation and action samples and condition the policy. (c): at
    inference, a schedule selects action offsets, and the executed action
    chunk determines the observation spacing for the next policy cycle.}
    \label{fig:method-overview}
\end{figure*}

For direction $\kappa\in\{+,-\}$, sample a trajectory $\tau$ from
$\widetilde{\mathcal{D}}^\kappa$, a valid anchor $t$, and a schedule $\eta$.
We construct $h_t^\eta$ and $A_t^\eta$ using \Cref{eq:temporal-window}
and their condition $c_t^\eta$ using \Cref{eq:temporal-condition}.
Using the conditional denoising loss defined in
\Cref{eq:bg-dp3-objective}, the training objective is
\begin{equation}
    \mathcal{L}(\theta)
    =\mathbb{E}_{\tau,t,\eta}
      \left[
        \mathcal{L}_{\mathrm{DP3}}
        (\theta;A_t^{\eta},c_t^{\eta})
      \right].
    \label{eq:combined-training-objective}
\end{equation}
This applies the conditional denoising objective in
\Cref{eq:bg-dp3-objective} to the augmented windows sampled under the
imitation-learning dataset. 
At inference, the scheduling rule $\sigma$
selects $\boldsymbol{\delta}^{a}_q$, which determines the future action
indices through
\Cref{eq:inference-action-schedule}. We obtain the observation offsets
from the preceding chunk using
\Cref{eq:inference-observation-offsets}. The resulting schedule conditions
the next action prediction together with the current observation history.
The appendix pseudocode in \Cref{alg:combined-method} summarizes both stages, and
\Cref{fig:method-overview} shows the method overview.

%% file: 5_Experimental_Results.tex
\section{Experimental Results}
\label{sec:experimental_results}

We evaluate \methodname{} on paired bimanual manipulation tasks from the RoboTwin
benchmark \citep{Mu_2025_CVPR,chen2025robotwin}. We first describe the benchmark,
training protocol, and evaluation metric in \Cref{sec:exp-setup}.
\Cref{sec:exp-main} compares \methodname{}
with DP3 at the same and twice the demonstration budget.
\Cref{sec:exp-ablation} examines action-offset selection at inference,
randomized observation offsets during training, and the individual effects
of trajectory reversal and temporal-stride augmentation.

\subsection{Experimental Setup}
\label{sec:exp-setup}

We use eight forward tasks from RoboTwin 2.0 and independently implement one
reverse counterpart per task, yielding eight pairs (16 directions; see
\Cref{sec:app-task-suite}). Each pair has expert datasets
$\mathcal{D}^{+}$ and $\mathcal{D}^{-}$; the former contains forward
demonstrations, while the latter is collected in our reverse environment with
our expert policy. Offline reversal augments each dataset with the reversed
trajectories from its partner, and these synthetic samples do not count toward
the data budget. DP3 is the baseline, with fixed observation and action
interfaces across methods. All models train for 3,000 epochs and are evaluated
for 100 episodes per task; we report per-task and mean success. Settings are
listed in \Cref{tab:app-hyperparameters}. Unless noted otherwise,
temporal-stride/observation-offset methods use the \texttt{rand} schedule,
which samples a gap from $\{1,2,3\}$ per action chunk and supplies the
corresponding observation offsets.

\subsection{Main Results}
\label{sec:exp-main}

\paragraph{Evaluation Success Rate}

\Cref{tab:main-results} compares the baseline and \methodname{} at two
collection budgets. Our goal is to test whether $\methodname{}_n$ can improve
over the same-budget baseline $B_n$ while approaching the performance of the
larger-data reference $B_{2n}$. Across the full task suite, the resulting
relationship is $B_n < \methodname{}_n \approx B_{2n}$. In practical terms,
\methodname{} can reach the performance of a policy trained with $2n$ native
demonstrations while collecting only $n$ trajectories and deriving additional
reversed augmentation trajectories offline. For a paired task, demonstrations
from one direction can therefore provide additional supervision for the
opposite direction without another collection run. We count only independently
collected demonstrations toward the data budget; reversed augmentation
trajectories are generated offline.

\begin{table*}[t]
    \centering
    \caption{Main results on all 16 paired tasks at 3,000 training epochs.
    $B_n$ is DP3 trained on $n$ native trajectories. $\methodname{}_n$ is DP3
    trained on the same $n$ native trajectories and their offline time-reversed
    counterparts, with temporal-stride augmentation and observation-offset
    conditioning. The subscript counts independently collected trajectories;
    $B_{2n}$ is DP3 trained on $2n$ native trajectories. Bold text with blue
    shading marks the best result at each current budget; $B_{100}$ is a
    reference upper bound and is not highlighted. Values are success rates (\%).
    Higher is better.}
    \label{tab:main-results}
    \setlength{\tabcolsep}{4pt}
    \begin{tabular}{lccccc}
        \toprule
        Task & $B_{25}$ & $\methodname{}_{25}$ & $B_{50}$ & $\methodname{}_{50}$ & $B_{100}$ \\
        \midrule
        Place Fan            & 23 & \bestresult{43} & 34 & \bestresult{55} & 65 \\
        Remove Fan           & 87 & \bestresult{96} & 92 & \bestresult{99} & 99 \\
        Place Object Stand   & 43 & \bestresult{58} & 61 & \bestresult{79} & 80 \\
        Remove Object Stand  & 64 & \bestresult{75} & \bestresult{82} & 76 & 87 \\
        Place Empty Cup      & 47 & \bestresult{77} & 77 & \bestresult{91} & 93 \\
        Remove Empty Cup     & 91 & \bestresult{97} & 90 & \bestresult{100} & 100 \\
        Hang Mug             & 19 & \bestresult{25} & 24 & \bestresult{36} & 31 \\
        Unhang Mug           & \bestresult{79} & 72 & \bestresult{96} & 85 & 94 \\
        Rotate QR Code       & 55 & \bestresult{80} & 69 & \bestresult{88} & 75 \\
        Rotate QR Code Back  & 99 & \bestresult{100} & \bestresult{100} & 99 & 98 \\
        Handover Block       & 86 & \bestresult{92} & 89 & \bestresult{96} & 97 \\
        Handover Block Back  & 99 & \bestresult{100} & 97 & \bestresult{99} & 99 \\
        Open Laptop          & 70 & \bestresult{93} & 85 & \bestresult{97} & 86 \\
        Close Laptop         & \bestresult{73} & 69 & 88 & \bestresult{92} & 88 \\
        Open Microwave       & 72 & \bestresult{96} & 80 & \bestresult{98} & 92 \\
        Close Microwave      & 73 & \bestresult{82} & 73 & \bestresult{74} & 83 \\
        \midrule
        Average(\%)          & 67.5 & \bestresult{78.4} & 77.3 & \bestresult{85.2} & 85.4 \\
        \bottomrule
    \end{tabular}
\end{table*}

\paragraph{Execution-Step Analysis}

Success rate alone does not reveal the execution cost of successful episodes.
We report both policy actions and simulator physics steps until success.
Physics steps count each simulator update, including the low-level
interpolation used to execute an action. The simulator advances by $1/250$ s
per physics step, or 4 ms, so dividing the count by 250 gives the simulated
duration of an episode. This duration is a closer proxy for robot execution
time than the policy-action count, although it does not include policy
inference or other system overhead and is not a hardware timing measurement.
For each task, we average over successful episodes and then average the task
means across the sixteen tasks. In these units, the mean simulated duration is
57.0 s for $B_{25}$ and 37.3 s for $\mathrm{\methodname{}}_{25}$, and 52.9 s for
$B_{50}$ and 36.3 s for $\mathrm{\methodname{}}_{50}$; $B_{100}$ averages 51.5 s.
Thus, \methodname{} reduces this duration by 19.7 s and 16.6 s at the two matched
budgets, respectively. \Cref{tab:execution-step-summary} reports average
policy-action and physics-step counts; per-task results are provided in
\Cref{tab:baseline-resta-policy-steps,tab:baseline-resta-physics-steps}.

\begin{table}[t]
    \centering
    \caption{Average execution steps over the sixteen paired tasks.
    Each task mean uses all successful episodes. \methodname{} uses the
    \texttt{rand} offset schedule. Lower is better.}
    \label{tab:execution-step-summary}
    \setlength{\tabcolsep}{4pt}
    \begin{tabular}{lccccc}
        \toprule
        Measure & $B_{25}$ & $\mathrm{\methodname{}}_{25}$ & $B_{50}$
        & $\mathrm{\methodname{}}_{50}$ & $B_{100}$ \\
        \midrule
        Policy actions & 217.8 & \bestresult{100.3} & 209.2 & \bestresult{99.3} & 196.4 \\
        Physics steps & 14251.3 & \bestresult{9337.0} & 13223.3 & \bestresult{9063.5} & 12879.4 \\
        \bottomrule
    \end{tabular}
\end{table}

\subsection{Ablation Study}
\label{sec:exp-ablation}

We examine how to select action offsets at inference, whether to randomize
observation offsets during training, and how each augmentation contributes
at different demonstration budgets.

\paragraph{Action-offset Selection}

At $n=50$, we evaluate the same TSAO checkpoint under seven inference
schedules. The fixed schedules \texttt{offset1}, \texttt{offset2}, and
\texttt{offset3} use action gaps of one, two, and three, respectively.
The \texttt{rand} schedule samples one gap per chunk and applies it to every
target, whereas \texttt{rand\_vector} samples each gap independently.
The voting schedules \texttt{vote3} and \texttt{vote11} select among candidate
action chunks. This ablation provides the full analysis of action-schedule
selection; definitions of all seven inference schedules are provided in
\Cref{sec:app-schedule-selection}.
At $n=50$, \texttt{rand} has the highest average success among the seven
evaluated schedules (83.2\%), so we use it for \methodname{} and the
TSA--TSAO comparison. Full per-task schedule results are in
\Cref{tab:app-schedule-selection}.

\paragraph{Randomized Observation Offsets}

We compare temporal-stride augmentation with consecutive training observations
(TSA) against its extension with randomized observation offsets and the
corresponding conditioning (TSAO). Both variants randomize action gaps during
training and use \texttt{rand} at inference. TSAO exposes the policy to
observation histories with different temporal spacings, matching the variation
induced by previously executed action chunks.
TSAO improves over TSA at $n=25$ (74.6\% vs. 72.4\%) and at $n=50$
(83.2\% vs. 80.8\%), supporting the benefit of training on varied
observation histories and conditioning on their offsets. The full per-task
comparison is provided in \Cref{tab:app-tsa-obs-offset}.

\paragraph{Effects of the Proposed Techniques}

We compare trajectory reversal, temporal-stride augmentation, and their
combination at matched collection budgets. $\mathrm{TR}_n$ uses $n$ native
and $n$ time-reversed trajectories, whereas $\mathrm{TSAO}_n$ uses $n$ native
trajectories with temporal-stride augmentation and observation-offset
conditioning, evaluated with the selected \texttt{rand} schedule.
$\methodname{}_n$ combines trajectory reversal with this temporal-stride and
observation-offset augmentation. The full per-task results and averages are
provided in \Cref{tab:app-data-efficiency}. $\mathrm{TSAO}_n$ is the
stronger aggregate component under the randomized inference schedule,
consistently improving average success across all evaluated collection
settings. $\mathrm{TR}_n$ can be more helpful on individual
paired tasks whose demonstrations transfer well across directions. Combining
the two gives the strongest current-budget result and approaches the larger-data
reference without additional independent collection.

%% file: 6_Conclusion.tex
\section{Conclusion and Limitations}
\label{sec:conclusion}

\methodname{} combines offline paired-task reversal, temporal strides, and
temporal offset conditioning. Across 16 simulated RoboTwin tasks, it raises
success from 67.5\% to 78.4\% at 25 demonstrations and from 77.3\% to 85.2\%
at 50, nearing DP3's 85.4\% at 100 while shortening episodes
(\Cref{tab:main-results,tab:execution-step-summary}). Simulation-only tests and
infeasible reversals limit our evidence; future work includes automatic
task-pair discovery.

%% file: 7_AI_statement.tex
\clearpage
\section*{AI use statement}

We used generative AI tools to assist with drafting and revising selected
manuscript text; writing code and running experiments; generating tables from
the experimental results; and designing and editing the teaser figure,
including its figure-production code. AI tools were not used to fabricate or
alter experimental measurements. We reviewed AI-assisted code, experiment
outputs, tables, text, and visual materials, and checked reported values
against the experiment results. The authors take responsibility for the final
content of this work, including text, claims, code, tables, and figures
produced with AI assistance.

%% file: appendix.tex
\appendix

\section{Additional Related Work}
\label{sec:app-related-work}

\paragraph{Action-Chunk Policies for Imitation Learning}

Action-chunk policies predict multiple future targets for receding-horizon
execution. Action Chunking with Transformers (ACT) \pw{\citep{zhao2023act}}
predicts chunks from visual observations, while Diffusion Policy
\citep{chi2023diffusionpolicy} models action sequences with conditional
denoising diffusion. 3D Diffusion Policy (DP3)
\pw{\citep{ze2024dp3}} conditions this formulation on compact point-cloud
and proprioceptive features. We use DP3 as our backbone for its geometric and
proprioceptive inputs, keeping its architecture fixed to isolate our data
augmentations: reversing demonstrations across paired task directions and
sampling observation--action windows at varied temporal strides and offsets.

Other methods improve demonstration efficiency or robustness by learning from
a single demonstration \citep{johns2021coarse}, synthesizing trajectories
with GANs \citep{wang2021fulltrajectorygan}, or perturbing MPC-generated data
\citep{tagliabue2024tubeguided}. DrQ-v2 and DrAC augment visual-control data
or select augmentations automatically in reinforcement learning
\citep{yarats2021drqv2,raileanu2020drac}. These methods alter observations,
state--action coverage, or trajectory content; our method instead changes the
temporal indices used to form imitation-learning windows.

\paragraph{Temporal Frequency and Multi-Rate Action Generation}

Temporal adaptation has been studied for robot-assisted rehabilitation
imitation \citep{dong2025temporalrehab}, time-driven demonstrations for robot
anomaly recovery \citep{wu2021timedrivenrecovery}, and temporal abstraction in
offline RL \citep{ayyagari2024temporalabstraction}. HiPolicy
\pw{\citep{zhang2026hipolicy}} predicts chunks at multiple frequencies,
fuses frequency-aligned histories, and selects execution frequency using an
entropy-based rule. FLASH
\pw{\citep{bai2026flash}} combines sparse temporal sampling with a continuous
Legendre-polynomial action representation for long-horizon flow matching;
Proleptic Temporal Ensemble \pw{\citep{park2025proleptic}} changes
inference-time temporal aggregation to accelerate execution. These methods
address different tasks or alter policy representations and inference. Our
generic data-level approach changes training-window sampling and supplies the
temporal offsets to an existing action-chunk policy, without frequency-specific
branches or an entropy-based selector; it improves results under fixed and
randomized schedules.

\section{Additional Experimental Details}
\label{sec:appendix-details}

This section reports the DP3 configuration and the temporal-offset settings
used in our experiments. All methods use DP3 as the policy backbone; the
table separates its model and optimization settings from offsets used by
augmented variants.

\begin{table*}[t]
    \centering
    \caption{Model, optimization, and temporal-offset settings. The settings
    are shared across data budgets; augmentation is enabled only for the
    corresponding method variants.}
    \label{tab:app-hyperparameters}
    \begin{tabular}{p{0.20\textwidth}p{0.74\textwidth}}
        \toprule
        Component & Setting \\
        \midrule
        Inputs & 1,024 point-cloud points; 14-D robot state and action \\
        Chunking & 3 observation steps; horizon 8; predict and execute 6 action targets \\
        DP3 backbone & FiLM-conditioned U-Net with channel widths $512$, $1024$, and $2048$; 128-D point-cloud feature \\
        Diffusion & 100 training steps with the squared-cosine schedule; 10 DDIM inference steps \\
        Optimizer & AdamW; learning rate $10^{-4}$; weight decay $10^{-6}$; betas $(0.95,0.999)$ \\
        Learning-rate schedule & Cosine decay with 500 warm-up steps \\
        Batching & Batch size 256; one gradient-accumulation step \\
        Training & 3,000 epochs; seed 0; 2\% validation split; EMA enabled with maximum decay 0.9999 \\
        Training action strides & For temporal-stride variants, independently sample each of the 6 action gaps uniformly from $\{1,2,3\}$ \\
        Training observation offsets & TSAO and \methodname{} independently sample 2 observation skip counts from $\{0,1,2\}$; TSA retains consecutive observations \\
        Offset conditioning & TSA conditions on 12 action-offset values (6 gaps and 6 cumulative gaps); TSAO and \methodname{} add the 2 observation skips for a 14-D input \\
        Inference schedule & The \texttt{rand} schedule samples one action gap uniformly from $\{1,2,3\}$ per chunk and applies it to all 6 targets; TSAO and \methodname{} derive the next observation spacing from the executed action gaps \\
        Evaluation & 100 episodes per task; report success rate and average task-level success \\
        \bottomrule
    \end{tabular}
\end{table*}

For completeness, \Cref{alg:combined-method} gives the combined training and
inference procedure.

\begin{algorithm}[t]
\caption{\methodname{} algorithm}
\label{alg:combined-method}
\begin{algorithmic}[1]
\Require Paired datasets $\mathcal{D}^{+},\mathcal{D}^{-}$, history length
$L$, action horizon $H\geq L$, stride set $\mathcal{B}$
\Procedure{Train}{$\kappa\in\{+,-\}$}
    \State Form $\widetilde{\mathcal{D}}^\kappa$ using
    \Cref{eq:augmented-paired-datasets}
    \While{training}
        \State Sample $\tau\in\widetilde{\mathcal{D}}^\kappa$ and valid anchor $t$
        \State Randomly sample $\boldsymbol{\delta}^{o}\in\mathcal{B}^{L-1}$ and
        $\boldsymbol{\delta}^{a}\in\mathcal{B}^{H}$ independently
        \State Construct $\eta$, $h_t^{\eta}$, and $A_t^{\eta}$ using
        \Cref{eq:stride-index-relations,eq:temporal-window}
        \State Compute $c_t^{\eta}$ using \Cref{eq:temporal-condition}
        \State Update $\theta$ with \Cref{eq:combined-training-objective}
    \EndWhile
\EndProcedure
\Procedure{Infer}{$\pi_\theta,\sigma$}
    \State Initialize history, its observation offsets, and $q\gets 0$
    \While{task execution continues}
        \State Select adjacent action offsets $\boldsymbol{\delta}^{a}_q\in\mathcal{B}^H$ using $\sigma$
        \State Compute target indices from $\boldsymbol{\delta}^{a}_q$ using
        \Cref{eq:inference-action-schedule}
        \State Set $\eta_q\gets[\boldsymbol{\delta}^{o}_q,\boldsymbol{\delta}^{a}_q]$
        \State Predict an action chunk conditioned on history and $\eta_q$
        \State Execute the chunk and retain the last $L$ observations
        \State $q\gets q+1$
        \State Update observation offsets using
        \Cref{eq:inference-observation-offsets}
    \EndWhile
\EndProcedure
\end{algorithmic}
\end{algorithm}

\subsection{Task Suite}
\label{sec:app-task-suite}

\begin{figure*}[p]
    \centering
    \begin{minipage}[t]{0.48\textwidth}
        \centering
        \includegraphics[width=\linewidth]{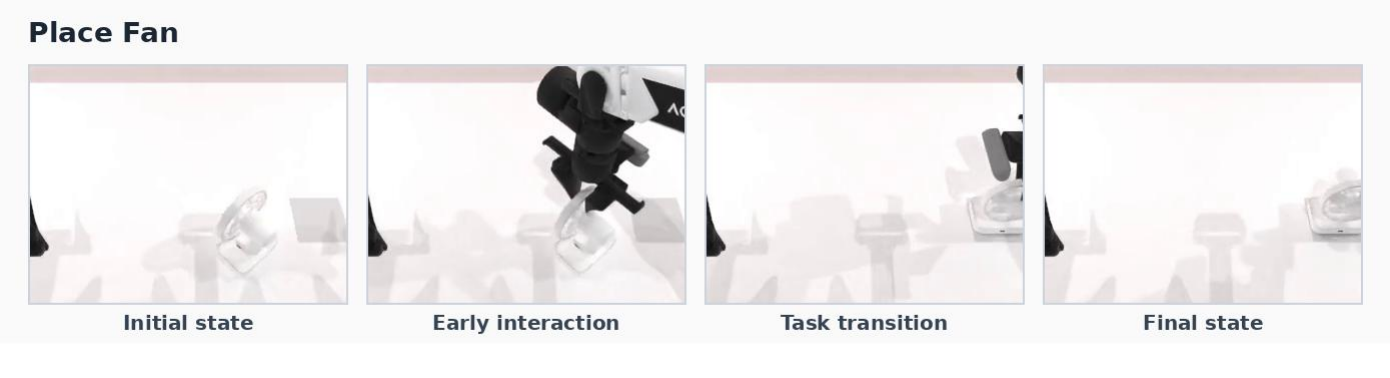}\par
        \scriptsize\textbf{Place Fan.} Place the fan at its target.
    \end{minipage}\hfill
    \begin{minipage}[t]{0.48\textwidth}
        \centering
        \includegraphics[width=\linewidth]{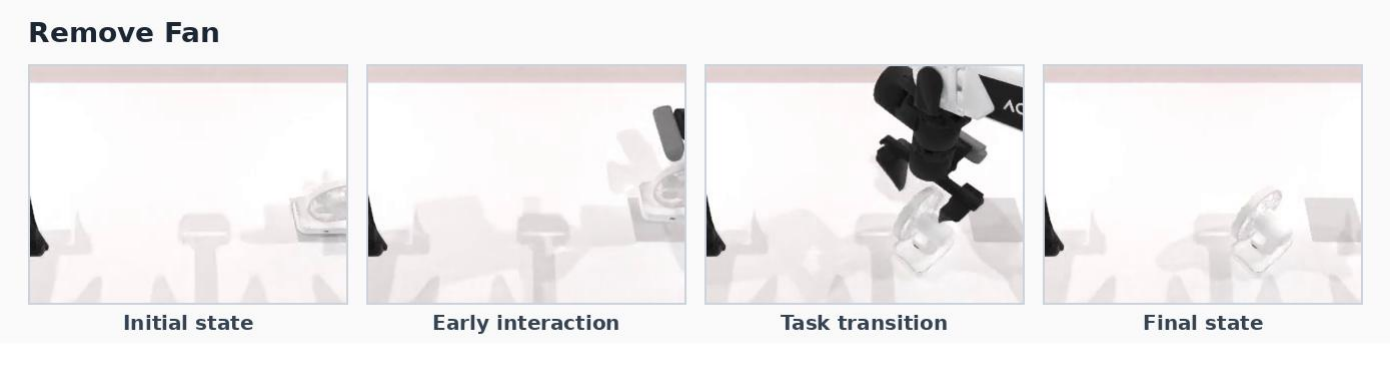}\par
        \scriptsize\textbf{Remove Fan.} Move the fan away from its target.
    \end{minipage}\par\vspace{2pt}
    \begin{minipage}[t]{0.48\textwidth}
        \centering
        \includegraphics[width=\linewidth]{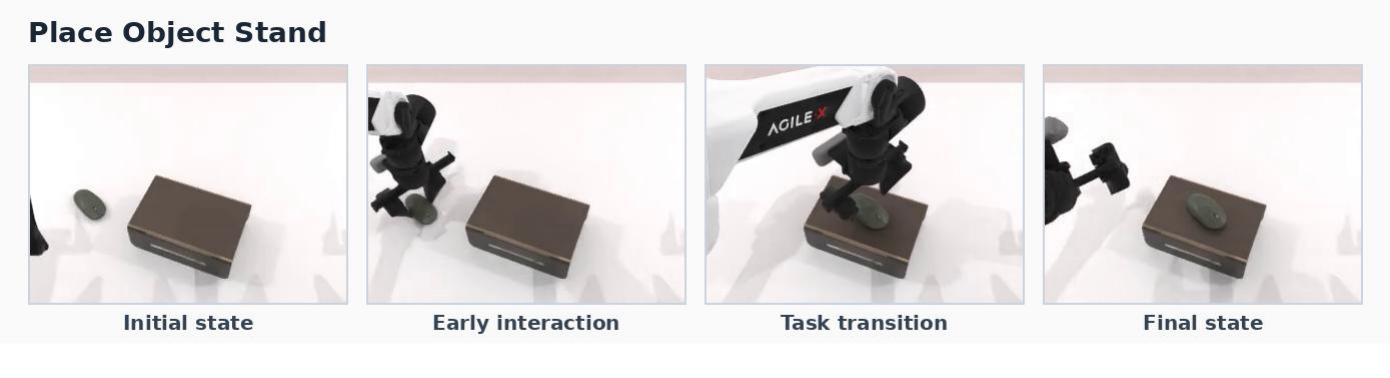}\par
        \scriptsize\textbf{Place Object Stand.} Place the stand at its target.
    \end{minipage}\hfill
    \begin{minipage}[t]{0.48\textwidth}
        \centering
        \includegraphics[width=\linewidth]{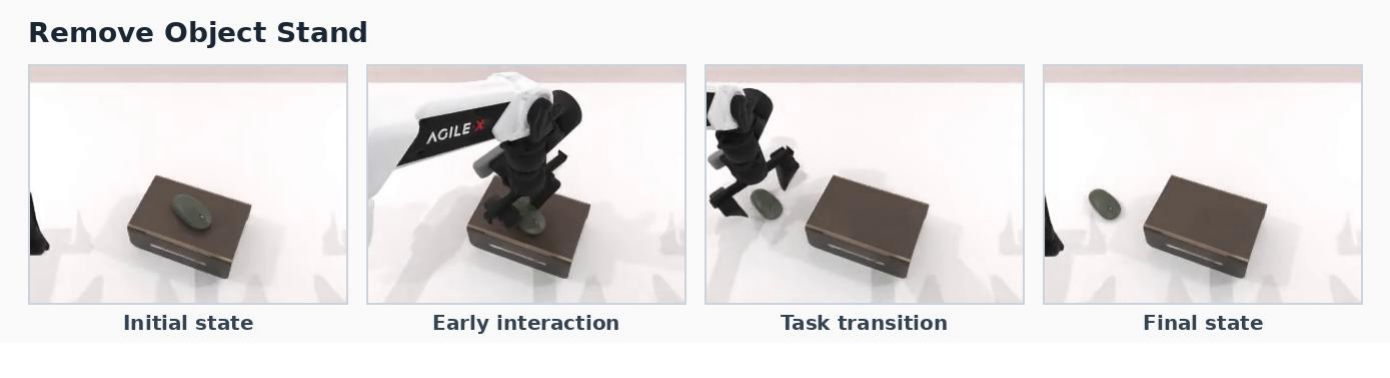}\par
        \scriptsize\textbf{Remove Object Stand.} Move the stand away from its target.
    \end{minipage}\par\vspace{2pt}
    \begin{minipage}[t]{0.48\textwidth}
        \centering
        \includegraphics[width=\linewidth]{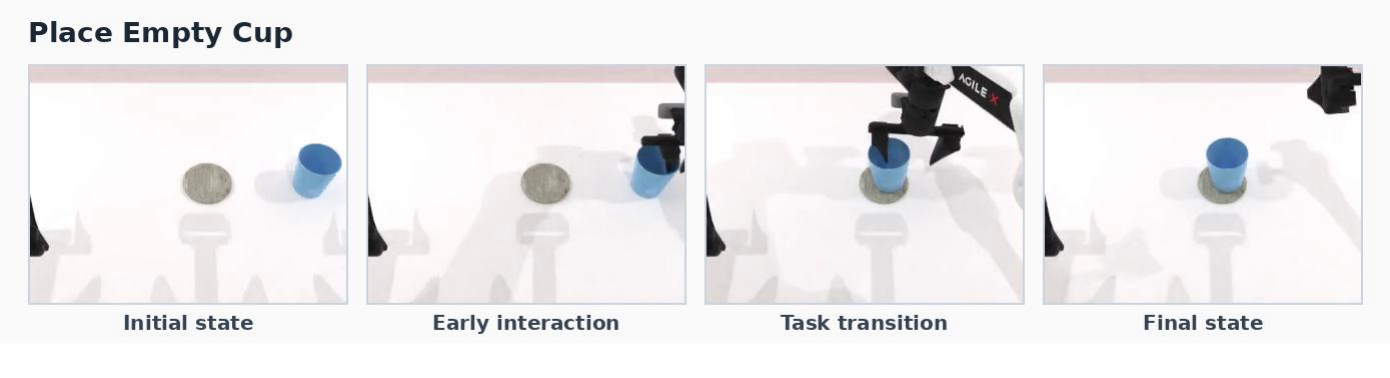}\par
        \scriptsize\textbf{Place Empty Cup.} Place the cup at its target.
    \end{minipage}\hfill
    \begin{minipage}[t]{0.48\textwidth}
        \centering
        \includegraphics[width=\linewidth]{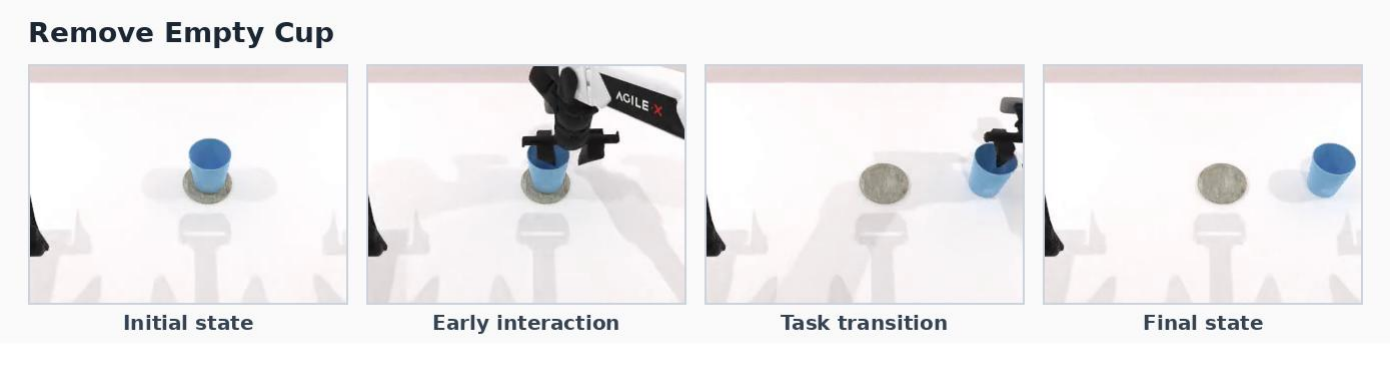}\par
        \scriptsize\textbf{Remove Empty Cup.} Move the cup away from its target.
    \end{minipage}\par\vspace{2pt}
    \begin{minipage}[t]{0.48\textwidth}
        \centering
        \includegraphics[width=\linewidth]{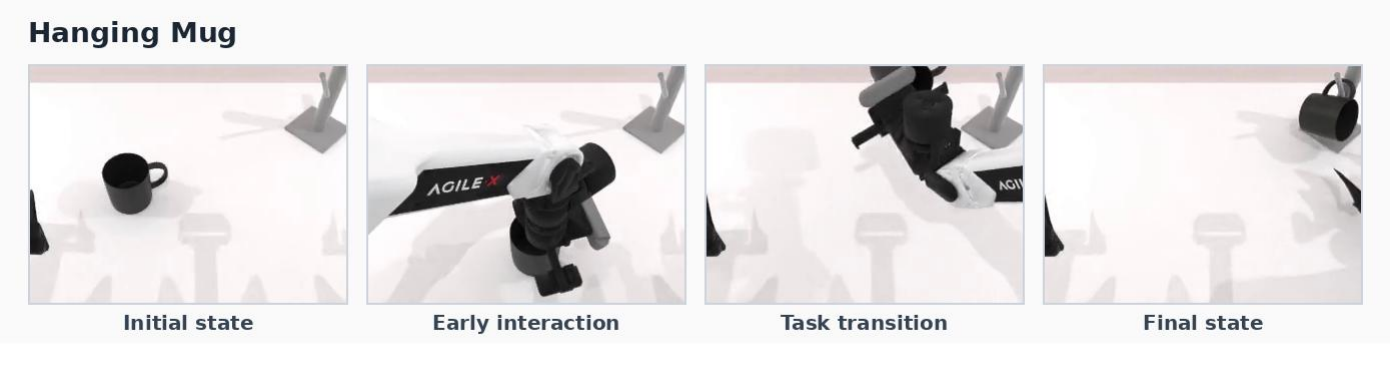}\par
        \scriptsize\textbf{Hang Mug.} Hang the mug on the fixture.
    \end{minipage}\hfill
    \begin{minipage}[t]{0.48\textwidth}
        \centering
        \includegraphics[width=\linewidth]{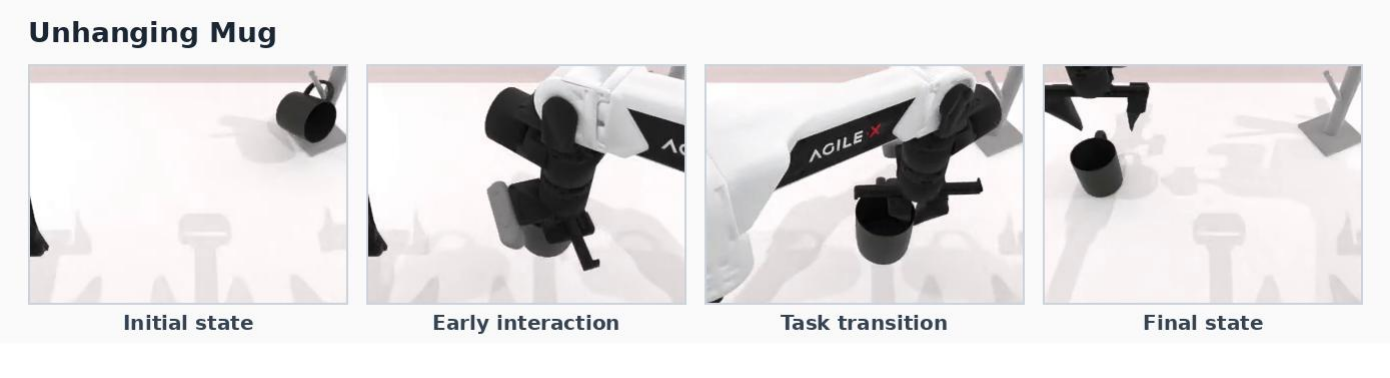}\par
        \scriptsize\textbf{Unhang Mug.} Remove the mug from the fixture.
    \end{minipage}\par\vspace{2pt}
    \begin{minipage}[t]{0.48\textwidth}
        \centering
        \includegraphics[width=\linewidth]{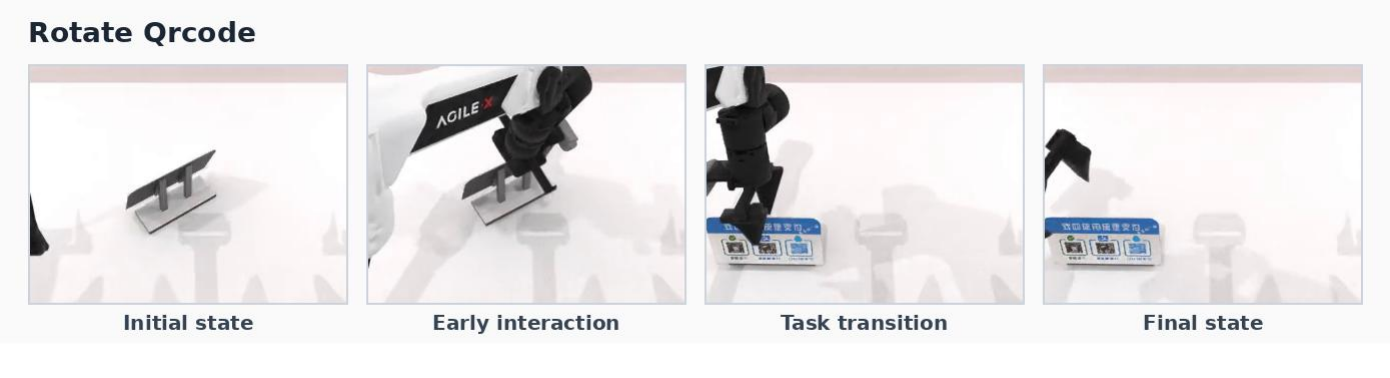}\par
        \scriptsize\textbf{Rotate QR Code.} Rotate the board to its target orientation.
    \end{minipage}\hfill
    \begin{minipage}[t]{0.48\textwidth}
        \centering
        \includegraphics[width=\linewidth]{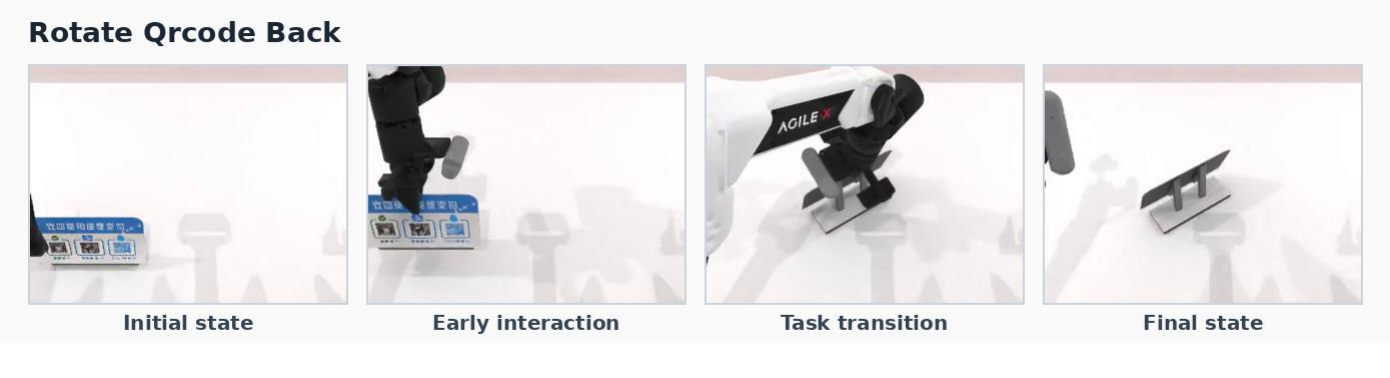}\par
        \scriptsize\textbf{Rotate QR Code Back.} Return the board toward its initial orientation.
    \end{minipage}\par\vspace{2pt}
    \begin{minipage}[t]{0.48\textwidth}
        \centering
        \includegraphics[width=\linewidth]{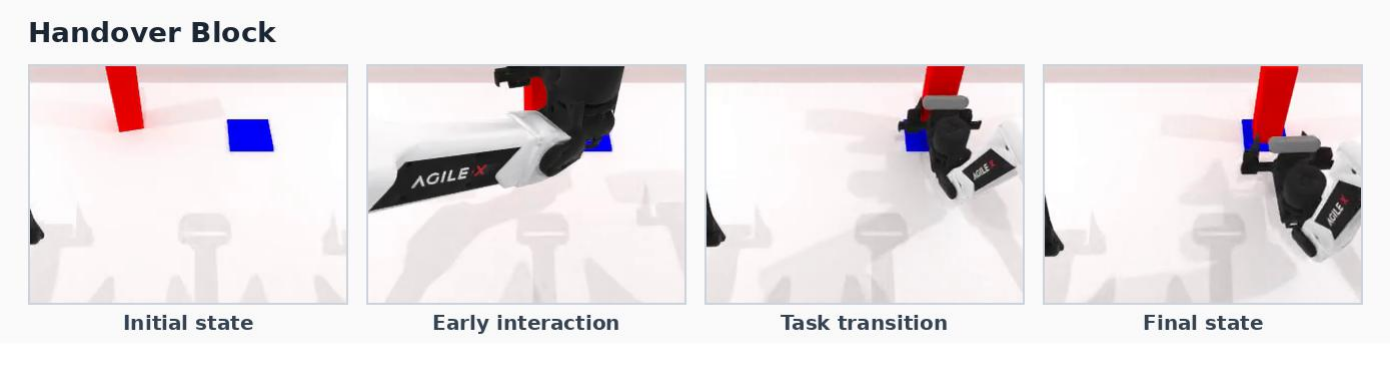}\par
        \scriptsize\textbf{Handover Block.} Transfer the block between grippers.
    \end{minipage}\hfill
    \begin{minipage}[t]{0.48\textwidth}
        \centering
        \includegraphics[width=\linewidth]{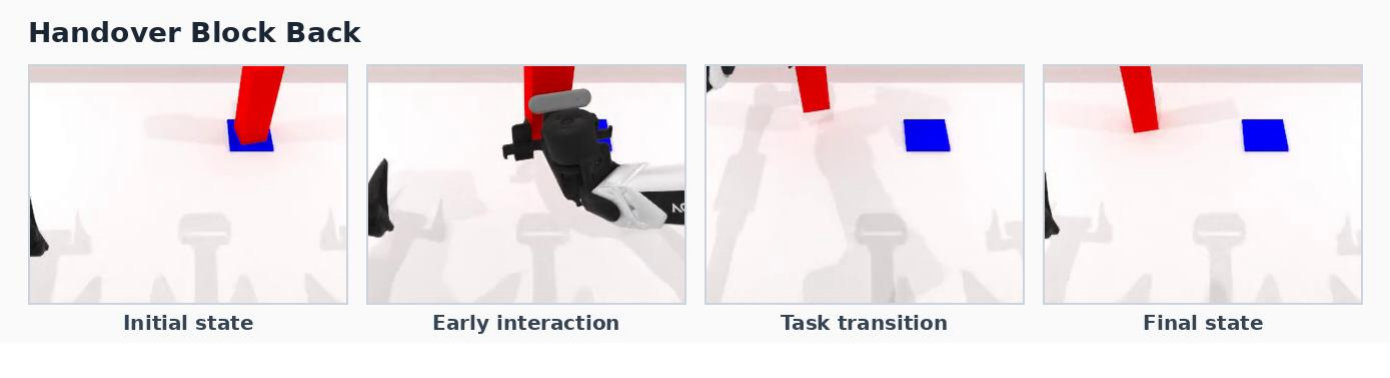}\par
        \scriptsize\textbf{Handover Block Back.} Transfer the block in reverse.
    \end{minipage}\par\vspace{2pt}
    \begin{minipage}[t]{0.48\textwidth}
        \centering
        \includegraphics[width=\linewidth]{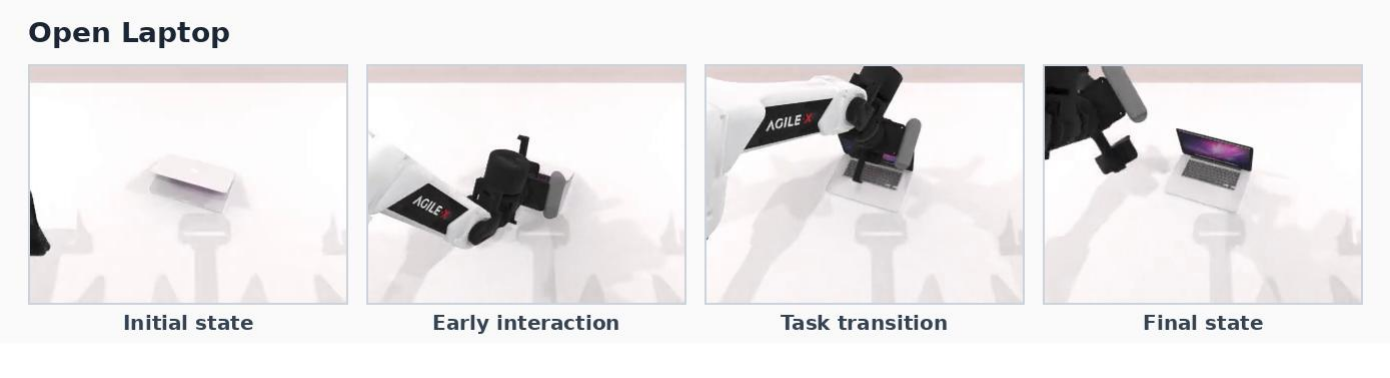}\par
        \scriptsize\textbf{Open Laptop.} Raise the laptop display.
    \end{minipage}\hfill
    \begin{minipage}[t]{0.48\textwidth}
        \centering
        \includegraphics[width=\linewidth]{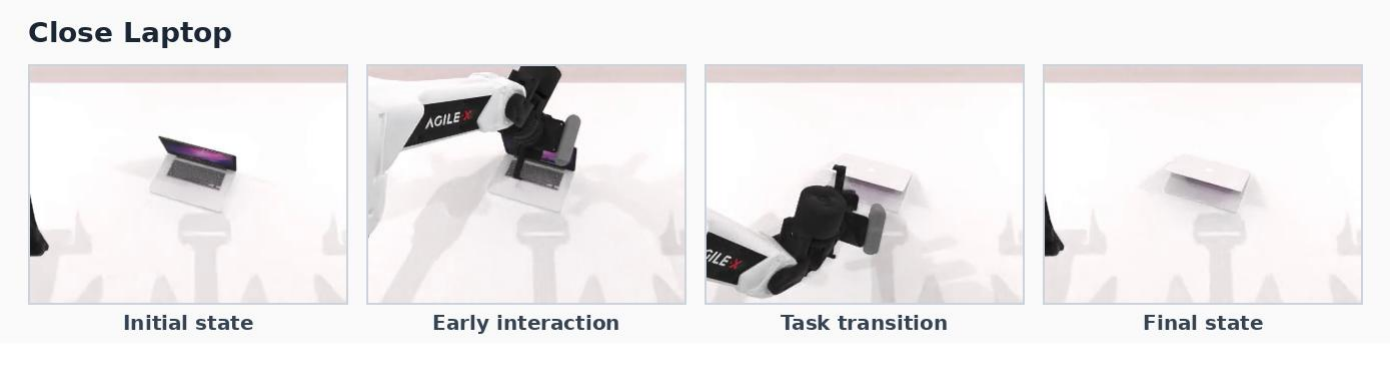}\par
        \scriptsize\textbf{Close Laptop.} Lower the laptop display.
    \end{minipage}\par\vspace{2pt}
    \begin{minipage}[t]{0.48\textwidth}
        \centering
        \includegraphics[width=\linewidth]{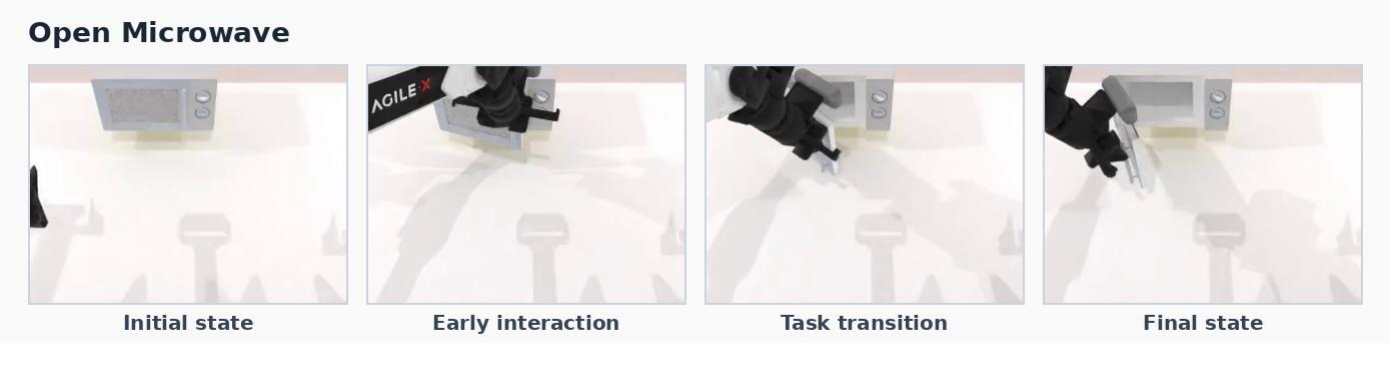}\par
        \scriptsize\textbf{Open Microwave.} Open the microwave door.
    \end{minipage}\hfill
    \begin{minipage}[t]{0.48\textwidth}
        \centering
        \includegraphics[width=\linewidth]{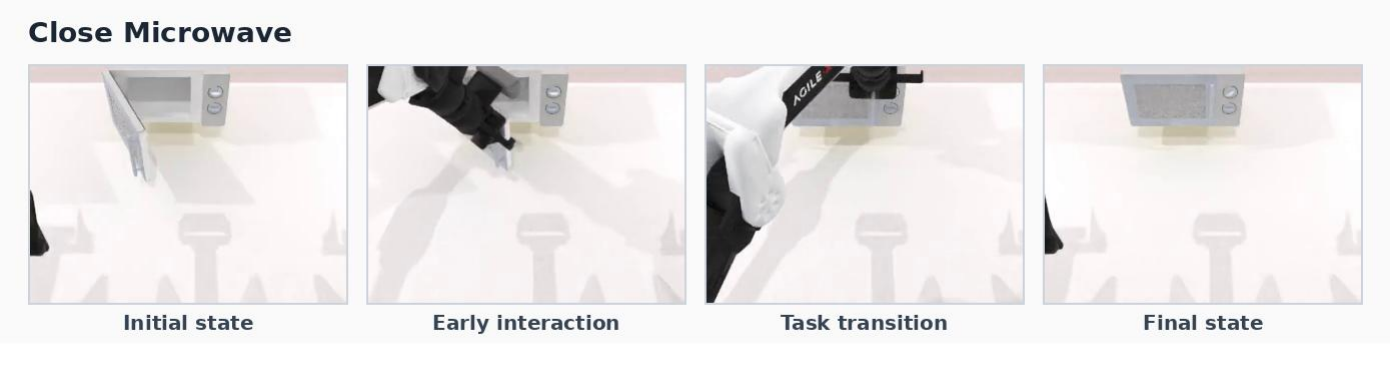}\par
        \scriptsize\textbf{Close Microwave.} Close the microwave door.
    \end{minipage}
    \caption{Four-frame expert demonstrations for all sixteen tasks, arranged
    as eight paired rows with the two task directions side by side.}
    \label{fig:app-task-demos}
\end{figure*}

The experimental suite contains sixteen bimanual manipulation tasks organized
as eight pairs. We select one forward task from RoboTwin 2.0 for each pair and
independently implement its reverse-task counterpart. For each pair, we assume
expert datasets $\mathcal{D}^{+}$ and $\mathcal{D}^{-}$; we create the
reverse-task environment and implement its expert policy to collect
$\mathcal{D}^{-}$. We then apply trajectory reversal offline in both
directions, using $\operatorname{Rev}(\mathcal{D}^{-})$ to augment
$\mathcal{D}^{+}$ and $\operatorname{Rev}(\mathcal{D}^{+})$ to augment
$\mathcal{D}^{-}$. Thus, trajectory reversal augments the independently
collected demonstrations but does not define or replace the reverse task
itself. \Cref{fig:app-task-demos}
shows expert demonstrations for all sixteen task directions; each sequence
contains four frames from the initial state to the final state. The task-pair
correspondences are listed in \Cref{tab:app-task-suite}.

\begin{table*}[t]
    \centering
    \caption{Forward tasks selected from RoboTwin 2.0 and their independently
    implemented reverse-task counterparts. Direction: not applicable; this
    table lists task correspondences.}
    \label{tab:app-task-suite}
    \begin{tabular}{lll}
        \toprule
        Task pair & Forward task & Reverse task \\
        \midrule
        Fan manipulation & Place Fan & Remove Fan \\
        Object-stand manipulation & Place Object Stand & Remove Object Stand \\
        Empty-cup manipulation & Place Empty Cup & Remove Empty Cup \\
        Mug manipulation & Hang Mug & Unhang Mug \\
        QR-code manipulation & Rotate QR Code & Rotate QR Code Back \\
        Block handover & Handover Block & Handover Block Back \\
        Laptop manipulation & Open Laptop & Close Laptop \\
        Microwave manipulation & Open Microwave & Close Microwave \\
        \bottomrule
    \end{tabular}
\end{table*}

\subsection{Effects of the Proposed Techniques}
\label{sec:app-data-efficiency}

\Cref{tab:app-data-efficiency} reports per-task and average results for the
data-efficiency comparison. The effects of the two augmentations vary across
tasks. At $n=25$, trajectory reversal improves Place Empty Cup
from 47\% to 83\%, whereas TSAO reaches 71\%. For Open Laptop, TSAO reaches
89\%, compared with 81\% for trajectory reversal and 70\% for the baseline.
At $n=50$, trajectory reversal improves Place Fan from 34\% to 58\%, but
reduces Unhang Mug from 96\% to 76\%. These task-level differences help
explain why reversal yields little aggregate gain at this budget.

\begin{table*}[t]
    \centering
    \caption{Per-task data-efficiency comparison at 3,000 training epochs.
    $B_n$ is DP3 trained with $n$ native trajectories per task.
    $\mathrm{TR}_n$ adds $n$ time-reversed trajectories, while
    $\mathrm{TSAO}_n$ uses only the $n$ native trajectories with
    temporal-stride augmentation and observation-offset conditioning.
    $\methodname{}_n$ combines trajectory reversal with this augmentation.
    TSAO and $\methodname{}_n$ use the \texttt{rand} inference schedule.
    $B_{100}$ is the larger-data reference and is not highlighted. Bold text
    with blue shading indicates the best result at each current budget.
    Values are success rates (\%). Higher is better.}
    \label{tab:app-data-efficiency}
    \setlength{\tabcolsep}{2pt}
    \begin{tabular}{lccccccccc}
        \toprule
        Task & $B_{25}$ & $\mathrm{TR}_{25}$ & $\mathrm{TSAO}_{25}$
        & $\methodname{}_{25}$ & $B_{50}$ & $\mathrm{TR}_{50}$
        & $\mathrm{TSAO}_{50}$ & $\methodname{}_{50}$ & $B_{100}$ \\
        \midrule
        Place Fan           & 23 & 34  & 30  & \bestresult{43} & 34  & \bestresult{58} & 38  & 55  & 65  \\
        Remove Fan          & 87 & 82  & \bestresult{100} & 96  & 92  & 85  & \bestresult{100} & 99  & 99  \\
        Place Object Stand  & 43 & \bestresult{61} & 50  & 58  & 61  & 72 & 67  & \bestresult{79}  & 80  \\
        Remove Object Stand & 64 & 67  & 68  & \bestresult{75} & 82  & 80  & \bestresult{85} & 76  & 87  \\
        Place Empty Cup     & 47 & \bestresult{83} & 71  & 77  & 77  & 74  & 85  & \bestresult{91} & 93  \\
        Remove Empty Cup    & 91 & 94  & 95  & \bestresult{97} & 90  & 98  & \bestresult{99} & \bestresult{100} & 100 \\
        Hang Mug            & 19 & \bestresult{27} & 21  & 25  & 24  & 28 & 28 & \bestresult{36}  & 31  \\
        Unhang Mug          & 79 & 77  & \bestresult{82} & 72  & \bestresult{96} & 76  & \bestresult{96} & 85  & 94  \\
        Rotate QR Code      & 55 & 55  & 65 & \bestresult{80}  & 69  & 78  & 79 & \bestresult{88}  & 75  \\
        Rotate QR Code Back & 99 & 98  & \bestresult{100} & \bestresult{100} & \bestresult{100} & \bestresult{100} & 99  & 99  & 98  \\
        Handover Block      & 86 & \bestresult{92} & 86  & \bestresult{92} & 89  & 94  & \bestresult{97} & 96  & 97  \\
        Handover Block Back & 99 & \bestresult{100} & 98  & \bestresult{100} & 97  & 97  & 98 & \bestresult{99}  & 99  \\
        Open Laptop         & 70 & 81  & 89  & \bestresult{93} & 85  & 81  & \bestresult{97} & \bestresult{97} & 86  \\
        Close Laptop        & 73 & 61  & \bestresult{79} & 69  & 88  & 73  & 93  & 92  & 88  \\
        Open Microwave      & 72 & 83  & 86  & \bestresult{96} & 80  & 73  & 95  & \bestresult{98} & 92  \\
        Close Microwave     & 73 & 77  & 74  & \bestresult{82} & 73  & 69  & \bestresult{75} & 74  & 83  \\
        \midrule
        Average(\%)         & 67.5 & 73.2 & 74.6 & \bestresult{78.4} & 77.3 & 77.2 & 83.2 & \bestresult{85.2} & 85.4 \\
        \bottomrule
    \end{tabular}
\end{table*}

\subsection{Temporal Schedule Selection}
\label{sec:app-schedule-selection}

The sweep compares seven schedulers for six-target chunks, with adjacent
action gaps from $\{1,2,3\}$. Their cumulative sums give target distances from
the anchor while keeping chunk length fixed. The fixed schedules
\texttt{offset1}, \texttt{offset2}, and \texttt{offset3} repeat gaps of one,
two, and three, spanning 6, 12, and 18 steps. \texttt{rand} samples one gap
per chunk, whereas \texttt{rand\_vector} samples each of the six gaps
independently.

The voting schedulers evaluate candidate gap vectors and execute the candidate
chunk closest to the mean predicted action chunk. \texttt{vote3} uses
\begin{equation}
\mathcal{S}_3^{\mathrm{vote}}=
\{(1,1,1,1,1,1),(2,2,2,2,2,2),(3,3,3,3,3,3)\},
\label{eq:app-vote3-schedules}
\end{equation}
and \texttt{vote11} expands this set with eight candidates:
\begin{equation}
\begin{aligned}
\mathcal{S}_{11}^{\mathrm{vote}}=\mathcal{S}_3^{\mathrm{vote}}\cup\{&
(1,2,3,1,2,3),(3,2,1,3,2,1),\\
&(1,3,2,1,3,2),(2,1,3,2,1,3),\\
&(1,1,2,2,3,3),(3,3,2,2,1,1),\\
&(1,2,2,3,3,2),(3,2,2,1,1,2)\}.
\end{aligned}
\label{eq:app-vote-schedules}
\end{equation}
The eight added \texttt{vote11} vectors test reordered three-gap cycles,
progressive coarsening or refinement, and nonmonotonic spacing. The first six
span 12 steps, matching \texttt{offset2}; the final two span 13 and 11 steps.
Together with the three constant vectors, they form a compact, non-exhaustive
subset of the $3^6$ possible schedules. The executed action gaps set the next
cycle's observation offsets. Comparing \texttt{vote3} and \texttt{vote11}
tests agreement-based selection, not task success directly; individual
\texttt{vote11} candidates were not ablated.

The preferred fixed gap varies by task: changing it from one to three raises
Open Laptop success from 86\% to 96\%, but lowers Close Microwave from 89\%
to 71\% (\Cref{tab:app-schedule-selection}). At $n=50$, \texttt{rand} has
the highest average success (83.2\%), versus 81.9\% for \texttt{rand\_vector},
and is used as the default (\Cref{tab:app-schedule-selection}). This favors
varying the scale between chunks while keeping regular spacing within each
chunk in this evaluation. \texttt{vote11} reaches 81.9\%, up from 81.6\% for
\texttt{vote3}, but remains below \texttt{rand}.

\begin{table*}[t]
    \centering
    \caption{Per-task success rates for the inference-schedule sweep in the
    $n=50$ setting, where $n$ is the number of native trajectories per task.
    The baseline is DP3; the remaining columns use TSAO and the indicated
    inference schedule. Bold and blue shading indicate the best result for
    each task or average across both blocks. Values are success rates (\%).
    Higher is better.}
    \label{tab:app-schedule-selection}
    \begin{tabular}{lcccc}
        \toprule
        Task & $B_{50}$ & \texttt{offset1} & \texttt{offset2} & \texttt{offset3} \\
        \midrule
        Place Fan            & 34 & 33 & 34 & 32 \\
        Remove Fan           & 92 & \bestresult{100} & \bestresult{100} & \bestresult{100} \\
        Place Object Stand   & 61 & 66 & \bestresult{67} & 63 \\
        Remove Object Stand  & 82 & 78 & 84 & 82 \\
        Place Empty Cup      & 77 & 79 & 85 & \bestresult{87} \\
        Remove Empty Cup     & 90 & 97 & 97 & 98 \\
        Hang Mug             & 24 & \bestresult{35} & 27 & 27 \\
        Unhang Mug           & \bestresult{96} & 96 & 96 & 95 \\
        Rotate QR Code       & 69 & 71 & 80 & 74 \\
        Rotate QR Code Back  & \bestresult{100} & \bestresult{100} & 99 & \bestresult{100} \\
        Handover Block       & 89 & 95 & 93 & 93 \\
        Handover Block Back  & 97 & 97 & 99 & \bestresult{100} \\
        Open Laptop          & 85 & 86 & 92 & 96 \\
        Close Laptop         & 88 & 92 & \bestresult{93} & 92 \\
        Open Microwave       & 80 & 80 & 96 & 96 \\
        Close Microwave      & 73 & \bestresult{89} & 73 & 71 \\
        \midrule
        Average(\%)          & 77.3 & 80.9 & 82.2 & 81.6 \\
        \bottomrule
    \end{tabular}

    \medskip
    \begin{tabular}{lcccc}
        \toprule
        Task & \texttt{rand} & \texttt{rand\_vector} & \texttt{vote3} & \texttt{vote11} \\
        \midrule
        Place Fan            & \bestresult{38} & 36 & 33 & 35 \\
        Remove Fan           & \bestresult{100} & \bestresult{100} & \bestresult{100} & \bestresult{100} \\
        Place Object Stand   & \bestresult{67} & \bestresult{67} & 66 & 65 \\
        Remove Object Stand  & \bestresult{85} & 81 & 81 & \bestresult{85} \\
        Place Empty Cup      & 85 & 85 & \bestresult{86} & 85 \\
        Remove Empty Cup     & \bestresult{99} & 97 & 98 & 98 \\
        Hang Mug             & 28 & \bestresult{30} & 21 & 26 \\
        Unhang Mug           & \bestresult{96} & \bestresult{96} & 95 & 90 \\
        Rotate QR Code       & 79 & 78 & 79 & \bestresult{83} \\
        Rotate QR Code Back  & 99 & 98 & 98 & 99 \\
        Handover Block       & \bestresult{97} & 93 & 94 & 92 \\
        Handover Block Back  & 98 & 99 & 98 & 98 \\
        Open Laptop          & \bestresult{97} & 95 & 92 & 92 \\
        Close Laptop         & \bestresult{93} & 91 & 92 & \bestresult{93} \\
        Open Microwave       & 95 & 94 & \bestresult{98} & 96 \\
        Close Microwave      & \bestresult{75} & 70 & \bestresult{75} & 73 \\
        \midrule
        Average(\%)          & \bestresult{83.2} & 81.9 & 81.6 & 81.9 \\
        \bottomrule
    \end{tabular}
\end{table*}

\subsection{Temporal-stride Ablation Details}
\label{sec:app-tsa-ablation}

The action schedule also determines the temporal spacing represented by
the next observation history, as specified in
\Cref{eq:inference-observation-offsets}. This motivates training with
varied observation spacings and conditioning on their offsets: the policy
must interpret histories produced by different preceding action schedules.
\Cref{tab:app-tsa-obs-offset} compares TSA and TSAO using the selected
\texttt{rand} schedule at both trajectory budgets. TSAO raises average
success from 72.4\% to 74.6\% at $n=25$ and from 80.8\% to 83.2\% at
$n=50$. The improvement is not uniform across tasks or budgets, supporting
the combined use of randomized observation spacings and offset conditioning
on average; this comparison does not isolate their individual effects.

\begin{table*}[t]
    \centering
    \caption{Per-task effect of temporal-stride augmentation and
    observation-offset conditioning at two trajectory budgets. The TSA and
    TSAO entries are the selected \texttt{rand} inference results from the
    corresponding budget-specific evaluations. For each
    $n\in\{25,50\}$, $B_n$ is the baseline, $\mathrm{TSA}_n$ uses
    temporal-stride augmentation with consecutive observations, and
    $\mathrm{TSAO}_n$ additionally conditions on observation offsets. Bold
    text with blue shading indicates the best result within each budget for
    each task or average across tasks. Values are success rates (\%). Higher is better.}
    \label{tab:app-tsa-obs-offset}
    \setlength{\tabcolsep}{4pt}
    \begin{tabular}{lcccccc}
        \toprule
        Task & $B_{25}$ & $\mathrm{TSA}_{25}$ & $\mathrm{TSAO}_{25}$
        & $B_{50}$ & $\mathrm{TSA}_{50}$ & $\mathrm{TSAO}_{50}$ \\
        \midrule
        Place Fan            & 23 & 18 & \bestresult{30} & 34 & 30 & \bestresult{38} \\
        Remove Fan           & 87 & \bestresult{100} & \bestresult{100} & 92 & \bestresult{100} & \bestresult{100} \\
        Place Object Stand   & 43 & \bestresult{55} & 50 & 61 & \bestresult{67} & \bestresult{67} \\
        Remove Object Stand  & 64 & 66 & \bestresult{68} & 82 & 81 & \bestresult{85} \\
        Place Empty Cup      & 47 & 69 & \bestresult{71} & 77 & 80 & \bestresult{85} \\
        Remove Empty Cup     & 91 & \bestresult{95} & \bestresult{95} & 90 & 97 & \bestresult{99} \\
        Hang Mug             & 19 & 17 & \bestresult{21} & 24 & 22 & \bestresult{28} \\
        Unhang Mug           & 79 & 79 & \bestresult{82} & \bestresult{96} & 95 & \bestresult{96} \\
        Rotate QR Code       & 55 & 61 & \bestresult{65} & 69 & 76 & \bestresult{79} \\
        Rotate QR Code Back  & 99 & 96 & \bestresult{100} & \bestresult{100} & 99 & 99 \\
        Handover Block       & \bestresult{86} & 85 & \bestresult{86} & 89 & 93 & \bestresult{97} \\
        Handover Block Back  & 99 & \bestresult{100} & 98 & 97 & \bestresult{99} & 98 \\
        Open Laptop          & 70 & 87 & \bestresult{89} & 85 & 90 & \bestresult{97} \\
        Close Laptop         & 73 & 67 & \bestresult{79} & 88 & 88 & \bestresult{93} \\
        Open Microwave       & 72 & \bestresult{87} & 86 & 80 & \bestresult{95} & \bestresult{95} \\
        Close Microwave      & 73 & \bestresult{76} & 74 & 73 & \bestresult{81} & 75 \\
        \midrule
        Average(\%)          & 67.5 & 72.4 & \bestresult{74.6} & 77.3 & 80.8 & \bestresult{83.2} \\
        \bottomrule
    \end{tabular}
\end{table*}

\subsection{Execution-Step Results}
\label{sec:app-execution-steps}

The following tables provide per-task execution-step measurements for the
baseline and \methodname{} checkpoints, including policy actions and simulator
physics steps. \methodname{} uses the \texttt{rand} schedule; values are means over
ten successful episodes per task.

\begin{table*}[t]
    \centering
    \caption{Mean policy actions per successful episode for baseline and \methodname{}
    checkpoints on the 16-task paired suite. \methodname{} uses the \texttt{rand}
    schedule; each task mean is based on all successful episodes. For each task
    and for the average across tasks, bold text with blue shading marks the lower value within
    each $n=25$ and $n=50$ baseline/\methodname{} pair. Lower is better.}
    \label{tab:baseline-resta-policy-steps}
    \begin{tabular}{lccccc}
        \toprule
        Task & $B_{25}$ & $\mathrm{\methodname{}}_{25}$ & $B_{50}$ & $\mathrm{\methodname{}}_{50}$ & $B_{100}$ \\
        \midrule
        Place Fan & 131.5 & \bestresult{71.9} & 143.8 & \bestresult{68.7} & 139.3 \\
        Remove Fan & 140.1 & \bestresult{70.4} & 135.9 & \bestresult{65.9} & 117.9 \\
        Place Object Stand & 132.9 & \bestresult{86.7} & 129.9 & \bestresult{64.9} & 130.9 \\
        Remove Object Stand & 159.2 & \bestresult{60.9} & 131.8 & \bestresult{63.2} & 112.7 \\
        Place Empty Cup & 149.6 & \bestresult{63.0} & 139.8 & \bestresult{58.0} & 144.0 \\
        Remove Empty Cup & 135.5 & \bestresult{63.4} & 142.1 & \bestresult{65.9} & 119.3 \\
        Hang Mug & 306.1 & \bestresult{143.2} & 333.0 & \bestresult{151.6} & 267.6 \\
        Unhang Mug & 335.5 & \bestresult{156.5} & 310.4 & \bestresult{172.7} & 289.9 \\
        Rotate QR Code & 163.7 & \bestresult{105.8} & 290.9 & \bestresult{71.7} & 130.3 \\
        Rotate QR Code Back & 151.3 & \bestresult{75.7} & 275.1 & \bestresult{68.9} & 133.7 \\
        Handover Block & 270.8 & \bestresult{128.3} & 144.1 & \bestresult{137.6} & 236.0 \\
        Handover Block Back & 295.0 & \bestresult{133.5} & 148.2 & \bestresult{132.0} & 255.5 \\
        Open Laptop & 114.0 & \bestresult{60.5} & 158.1 & \bestresult{67.5} & 117.7 \\
        Close Laptop & 172.4 & \bestresult{101.5} & 128.6 & \bestresult{84.5} & 149.1 \\
        Open Microwave & 624.6 & \bestresult{186.5} & 527.5 & \bestresult{215.7} & 588.8 \\
        Close Microwave & 203.2 & \bestresult{97.8} & 208.1 & \bestresult{100.1} & 209.3 \\
        \midrule
        Average & 217.8 & \bestresult{100.3} & 209.2 & \bestresult{99.3} & 196.4 \\
        \bottomrule
    \end{tabular}
\end{table*}

\begin{table*}[t]
    \centering
    \caption{Mean simulator physics steps per successful episode for baseline
    and \methodname{} checkpoints on the 16-task paired suite. Physics steps include
    low-level interpolation. \methodname{} uses the \texttt{rand} schedule; each task
    mean is based on all successful episodes. For each task and for the average
    across tasks, bold text with blue shading marks the lower value within each $n=25$ and
    $n=50$ baseline/\methodname{} pair. Lower is better.}
    \label{tab:baseline-resta-physics-steps}
    \begin{tabular}{lccccc}
        \toprule
        Task & $B_{25}$ & $\mathrm{\methodname{}}_{25}$ & $B_{50}$ & $\mathrm{\methodname{}}_{50}$ & $B_{100}$ \\
        \midrule
        Place Fan & 10734.4 & \bestresult{7908.4} & 10897.0 & \bestresult{7406.1} & 10494.5 \\
        Remove Fan & 11734.1 & \bestresult{5995.5} & 9287.3 & \bestresult{6789.6} & 9386.8 \\
        Place Object Stand & 10198.1 & \bestresult{8883.6} & 9574.5 & \bestresult{6259.4} & 9130.2 \\
        Remove Object Stand & 13009.2 & \bestresult{6246.6} & 9436.0 & \bestresult{6097.9} & 8730.4 \\
        Place Empty Cup & 9728.8 & \bestresult{5878.8} & 8935.4 & \bestresult{5460.3} & 8629.3 \\
        Remove Empty Cup & 9251.7 & \bestresult{5984.6} & 9447.8 & \bestresult{6354.0} & 8864.5 \\
        Hang Mug & 23211.5 & \bestresult{15110.3} & 23636.2 & \bestresult{15185.0} & 21226.0 \\
        Unhang Mug & 24263.4 & \bestresult{16084.7} & 23053.7 & \bestresult{18252.1} & 22690.2 \\
        Rotate QR Code & 14794.1 & \bestresult{12491.0} & 18528.5 & \bestresult{8239.4} & 11635.6 \\
        Rotate QR Code Back & 13078.9 & \bestresult{8950.1} & 18666.3 & \bestresult{8260.9} & 12341.6 \\
        Handover Block & 19377.1 & \bestresult{12323.4} & \bestresult{11801.3} & 13266.0 & 17585.8 \\
        Handover Block Back & 19259.8 & \bestresult{12444.9} & 12257.0 & \bestresult{12084.2} & 17652.0 \\
        Open Laptop & 7905.4 & \bestresult{5171.2} & 8361.7 & \bestresult{5915.5} & 7781.5 \\
        Close Laptop & 12146.0 & \bestresult{8367.9} & 9881.8 & \bestresult{7105.3} & 12231.5 \\
        Open Microwave & 18580.9 & \bestresult{10211.3} & 16921.9 & \bestresult{10920.0} & 16856.2 \\
        Close Microwave & 10746.9 & \bestresult{7339.9} & 10886.1 & \bestresult{7420.7} & 10833.7 \\
        \midrule
        Average & 14251.3 & \bestresult{9337.0} & 13223.3 & \bestresult{9063.5} & 12879.4 \\
        \bottomrule
    \end{tabular}
\end{table*}